\pdfoutput=1
\documentclass[preprint,journal]{vgtc}       

\ifpdf
  \pdfoutput=1\relax                   
  \usepackage{graphicx}                
  \DeclareGraphicsExtensions{.pdf,.png,.jpg,.jpeg} 
\else
  \usepackage{graphicx}                
  \DeclareGraphicsExtensions{.eps}     
\fi%

\graphicspath{{figures/}{pictures/}{images/}{./}} 

\usepackage{microtype}                 
\PassOptionsToPackage{warn}{textcomp}  
\usepackage{textcomp}                  
\usepackage{times}                     
\usepackage{cite}                      
\usepackage{tabu}                      
\usepackage{booktabs}                  
\usepackage{amsmath}
\usepackage{amssymb}
\usepackage{mathtools}
\usepackage{bm}

\usepackage{enumitem}

\usepackage[titletoc,toc,title]{appendix}
\usepackage[linesnumbered]{algorithm2e}
\SetAlCapSkip{0.5em}

\usepackage[usenames]{color}
\definecolor{blue}{RGB}{31,119,180}
\definecolor{orange}{RGB}{255, 127, 14}
\definecolor{green}{RGB}{44, 160, 44}
\definecolor{visblue}{RGB}{51,105,204}
\definecolor{visorange}{RGB}{255, 153, 0}
\definecolor{black}{RGB}{0, 0, 0}
\newcommand{\revise}[1]{\textcolor{black}{#1}}

\newcommand{\revision}[1]{\textcolor{black}{#1}}

\newcommand{\circletext}[1]{\raisebox{.5pt}{\textcircled{\raisebox{-.9pt} {#1}}}}

\makeatletter
\def\itemautorefname{\@gobble}
\makeatother

\newcommand{\bluestripe}{\raisebox{-1pt}{\includegraphics[height=1.2\fontcharht\font`\B]{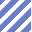}}}

\renewcommand{\manuscriptnotetxt}{}
\nocopyrightspace

\onlineid{1049}

\vgtccategory{Research}
\vgtcpapertype{algorithm/technique}

\title{RuleMatrix: Visualizing and Understanding Classifiers with Rules}

\author{Yao Ming, Huamin Qu, \textit{Member, IEEE}, and Enrico Bertini, \textit{Member, IEEE}}
\authorfooter{
\item
 Yao Ming is with Hong Kong University of Science and Technology. E-mail: ymingaa@ust.hk.
\item
 Huamin Qu is with Hong Kong University of Science and Technology. E-mail: huamin@cse.ust.hk.
\item
 Enrico Bertini is with New York University. Email: enrico.bertini@nyu.edu
}

\shortauthortitle{Biv \MakeLowercase{\textit{et al.}}: Global Illumination for Fun and Profit}

\abstract{With the growing adoption of machine learning techniques, there is a surge of research interest towards making machine learning systems more transparent and interpretable. Various visualizations have been developed to help model developers understand, diagnose, and refine machine learning models. However, a large number of potential but neglected users are the domain experts with little knowledge of machine learning but are expected to work with machine learning systems. In this paper, we present an interactive visualization technique to help users with little expertise in machine learning to understand, explore and validate predictive models. By viewing the model as a black box, we extract a standardized rule-based knowledge representation from its input-output behavior. We design RuleMatrix, a matrix-based visualization of rules to help users navigate and verify the rules and the black-box model. We evaluate the effectiveness of RuleMatrix via two use cases and a usability study.
} 


\CCScatlist{ 
 \CCScat{K.6.1}{Management of Computing and Information Systems}%
{Project and People Management}{Life Cycle};
 \CCScat{K.7.m}{The Computing Profession}{Miscellaneous}{Ethics}
}

\teaser{
  \centering
  \includegraphics[width=\linewidth]{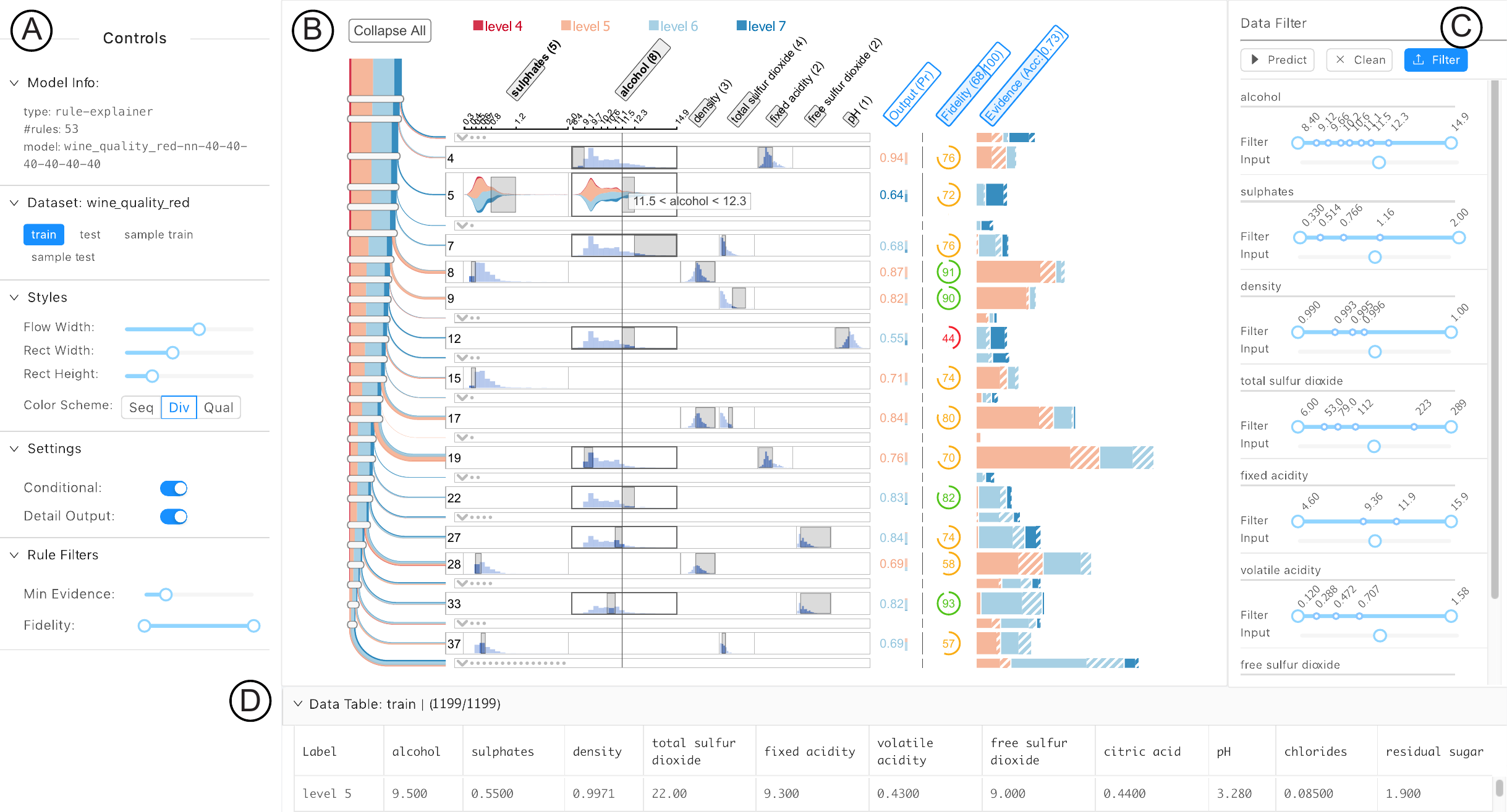}
  \caption{Understanding the behavior of a trained neural network using the explanatory visual interface of our proposed technique. The user uses the control panel (A) to specify the detail information to visualize (\eg, level of detail, rule filters). The rule-based explanatory representation is visualized as a matrix (B), where each row represents a rule, and each column is a feature used in the rules. The user can also filter the data or use a customized input in the data filter (C) and navigate the filtered dataset in the data table (D).}
	\label{fig:teaser}
}

\vgtcinsertpkg

\usepackage{color}
\newcommand{\etal}{\textit{et al}.}
\newcommand{\ie}{\textit{i}.\textit{e}.}
\newcommand{\eg}{\textit{e}.\textit{g}.}

\begin{document}


\firstsection{Introduction}

\maketitle


In this paper, we propose an interactive visualization technique for understanding and inspecting machine learning models. By constructing a rule-based interface from a given black box classifier, our method allows visual inspection of the reasoning logic of the model, as well as systematic exploration of the data used to train the model. 

\revision{With the recent advances in machine learning, there is increasing need for transparent and interpretable machine learning models} \cite{Ribeiro:2016:WIT, Caruana:2015:IMH, Krause:2017:WVD}. 
To avoid ambiguity, in this paper we define interpretability of a machine learning model as the ability to provide explanation for the reasoning of its prediction so that human users can understand.
Interpretability is a crucial requirement for machine learning models in applications where human users are expected to sufficiently understand and trust them. The need for interpretable machine learning has been addressed in medicine, finance, security \cite{Jan:2017:ASG} and many other domains where ethical treatment of data is required \cite{goodman2017a}. In a health care example given by Caruana \etal \cite{Caruana:2015:IMH}, logistic regression was chosen over neural networks due to interpretability concerns. Though the neural network achieved a significant higher receiver operating characteristic (ROC) score than the logistic regression, domain experts felt that it was too risky to deploy the neural network for decision making with real patients because of its lack of transparency. On the other hand, with logistic regression, though less accurate, the fitted parameters have relatively clearer meanings, which can facilitate the discovery \revision{of} problematic patterns in the dataset.

In the machine learning literature, trade-offs are often made between performance (\eg, accuracy) and interpretability. Models that are considered interpretable, such as logistic regression, k-nearest neighbors, and decision trees, often perform worse than models that are difficult to interpret, such as neural networks, support vector machines, and random forests. In scenarios where interpretability is required, the use of models with high performance is largely limited. There are two common strategies to strike a balance between performance and interpretability in machine learning. The first develops model simplification techniques (\eg, decision tree simplification \cite{Quinlan:1987:SDT}) that generate a sparser model without much performance degradation. The second aims to improve the performance by designing models with commonly-recognized interpretable structures (\eg, the linear relationships used by Generalized Additive Models (GAM) \cite{Caruana:2015:IMH} and decision rules employed by Bayesian Rule Lists \cite{Letham:2015:BRL}). However, simplification techniques are applicable to a certain type of model, which impedes their popularization. The newly emerged interpretable models, on the other hand, rarely retain a state-of-the-art performance along with interpretability.

Instead of struggling with the trade-offs, in this paper we explore the idea of introducing an extra explanatory interface between the human and the model to provide interpretability. The interface is created in two steps.
For a trained classification model, we first extract a rule list
 that approximates the original one using model induction. 
As a second step, we develop a visual interface to augment interpretability by enabling interactive exploration of details of the decision logic. The visual interface is crucial for numerous reasons.
Though rule-based models are commonly considered to be interpretable, their interpretability is largely weakened when the rule list is too long, or the composition of a rule is too complex. In addition, it is hard to identify how well the rules approximate the original model. 
The visual interface also enables the possibility to inspect the behavior of the model under a production environment, where the operators may not possess much knowledge about the underlying model.


In summary, the main contribution of this paper is a visual technique that helps domain experts understand and inspect classification models using rule-based explanation. We present two case studies and a user study to demonstrate the effectiveness of the proposed method. We also contribute a model induction algorithm that generates a rule list for any given classification model. 




\section{Related Work}
\revision{
Recent research has explored promising directions to make machine learning models explainable. By associating semantic information with a learned deep neural networks, researchers created visualizations that can explain the learned features of the model \cite{zeiler14,Simonyan:2014:saliency}. In another direction, a variety of algorithms has been developed to directly learn more interpretable and structured models, including generalized additive models \cite{debock2010gam} and decision rule lists \cite{Letham:2015:BRL,Yang:2016:sbrl}. Most related to our work, model-agnostic induction techniques \cite{Craven:1995:Trepan,Ribeiro:2016:WIT} have been used to generate explanations for any given machine learning model. 
}

\subsection{Model Induction}\label{related:model-induction}

Model induction is a technique that infers an approximate and interpretable model from any machine learning model. The inferred model can be a linear classifier \cite{Ribeiro:2016:WIT}, a decision tree \cite{Craven:1995:Trepan}, or a rule set \cite{Quinlan:1987:generateRule, Martens:2009:decomposeRule}. It has been increasingly applied to create human-comprehensible proxy models that help users make sense of the behavior of complex models, such as artificial neural networks and support vector machines (SVMs). One most desirable advantage of model induction is that it provides interpretability by treating any complex model as a black box without compromising the performance. 

There are mainly three types of methods to derive approximate models (often as rule sets) as summarized in related surveys \cite{Andrews:1995:survey, Augasta:2012:comparative}, namely, \textit{decompositional}, \textit{pedagogical} and \textit{eclectic}. Decompositional methods extract a simplified representation from specialized structures of a given model, \eg, the weights of a neural network, or the support vectors of an SVM, and thus only work for certain types of models. Pedagogical methods \revision{are often model-agnostic}, and learn a model that approximates the input-output behavior of the original one. Eclectic methods either combine the previous two, or have distinct differences from them. In this paper, we adopt pedagogical methods to obtain rule-based approximations due to their simplicity and generalizability.

However, as the complexity of the original model increases, model induction would also face trade-offs. We either learn a small and comprehensible model that fails to approximate the original model well, or we learn a well-approximated but large model (\eg, a decision tree with over 100 nodes) that can be hardly recognized as ``easy-to-understand''. In our work, we utilize visualization techniques to boost the interpretability while maintaining a good approximation quality.

\subsection{Visualization for Model Analysis}
Visualization has been increasingly used to support the understanding, diagnosis and refinement of machine learning models \revision{\cite{Liu:2017:survey,Abdul:2018:Trends}}. In pioneering work by Tzeng and Ma \cite{tzeng:2005:nn}, a node-linked visualization is used to understand and analyze a trained neural network's behavior in classifying volume and text data.

More recently, a number of visual analytics methods have been developed to support the analysis of complex deep neural networks \cite{Liu:2017:cnnvis, Rauber:2017:project, Strobelt:2018:lstmvis, Pezzotti:2018:deepeyes, Ming:2017:rnnvis, Bilal:2018:cnnHierarchy, Sacha:2017:HumanCenteredMachineLearning}. Liu \etal \cite{Liu:2017:cnnvis} used a hybrid visualization that embedded debugging information into the node-link diagram to help diagnose convolutional neural networks (CNNs). \revision{Alsallakh} \etal \cite{Bilal:2018:cnnHierarchy} stepped further to examine whether CNNs learn hierarchical representations from image data. Rauber \etal \cite{Rauber:2017:project} and Pezzotti \etal \cite{Pezzotti:2018:deepeyes} applied projection techniques to investigate the hidden activities of deep neural networks.
Ming \etal \cite{Ming:2017:rnnvis} developed a visual analytics method based on co-clustering to understand the hidden memories of recurrent neural networks (RNNs) in natural language processing tasks. Strobelt \etal \cite{Strobelt:2018:lstmvis} utilized parallel coordinates to help researchers validate hypotheses about the hidden state dynamics of RNNs. \revision{Sacha \etal \cite{Sacha:2017:HumanCenteredMachineLearning} introduced a human-centered visual analytics framework to incorporate human knowledge in the machine learning process.}

In the meantime, there are concrete demands in the industry to apply visualization to assist the development of machine learning systems \cite{Kahng:2018:ActiVis,Wongsuphasawat:2018:dataFlow}. 
Kahng \etal \cite{Kahng:2018:ActiVis} developed ActiVis, a visual system to support the exploration of industrial deep learning models in Facebook. Wongsuphasawat \etal \cite{Wongsuphasawat:2018:dataFlow} presented the TensorFlow Graph Visualizer, an integrated visualization tool to help  developers understand the complex structure of different machine learning architectures.

These methods have addressed the need for better visualization tools for machine learning researchers and developers. However, little attention has been paid to help domain experts (\eg, doctors and analysts) who have little or no knowledge of machine learning or deep learning to understand and exploit this powerful technology. Krause \etal \cite{Krause:2017:WVD} developed an overview-feature-item workflow to help explain machine learning models to domain experts operating a hospital. Such non-experts in machine learning are the major target users of our solution.

\subsection{Visualization of Rule-based Representations}
Rule-based models are composed of logical representations, that is, IF-THEN-ELSE statements which are pervasively used in programming languages. Typical representations of rule-based models include decision tables \cite{vanthienen:1994:decision}, decision trees \cite{breiman:1984:CART}, and rule sets or decision lists \cite{Rivest:1987:lists}. Among these representations, trees are hierarchical data that have been studied abundantly in visualization research. A gallery of tree visualization can be found on treevis.net \cite{schulz:2011:treevis}. Most related to our work, BaobabView \cite{elzen:2011:baobabview} \revision{uses a node-link data flow diagram to visualize the logic of decision trees, which inspired our design of data flow visualization in rule lists.}

However, there is little research on how visualization can help analyze decision tables and rule lists.  The lack of interest in visualizing decision tables and rule lists is partially due to the fact that they are not naturally graphical representations as trees. 
There is also no consensus that trees are the best visual representations for understanding rule-based models. A comprehensive empirical study conducted by Huysmans \etal \cite{Huysmans:2011:empirical} found that decision tables are the most effective representations, while other studies \cite{Allahyari:2011:UserorientedAO} disagrees. In a later position paper \cite{Freitas:2014:position}, Freitas summarized a few good properties rules and tables possess that trees do not. Also, all previous studies used pure texts to present rules. In our study, we provide a graphical representation of rule lists as an alternative for navigating and exploring proxy models.

\begin{figure*}[t]
 \centering 
 \includegraphics[width=\textwidth]{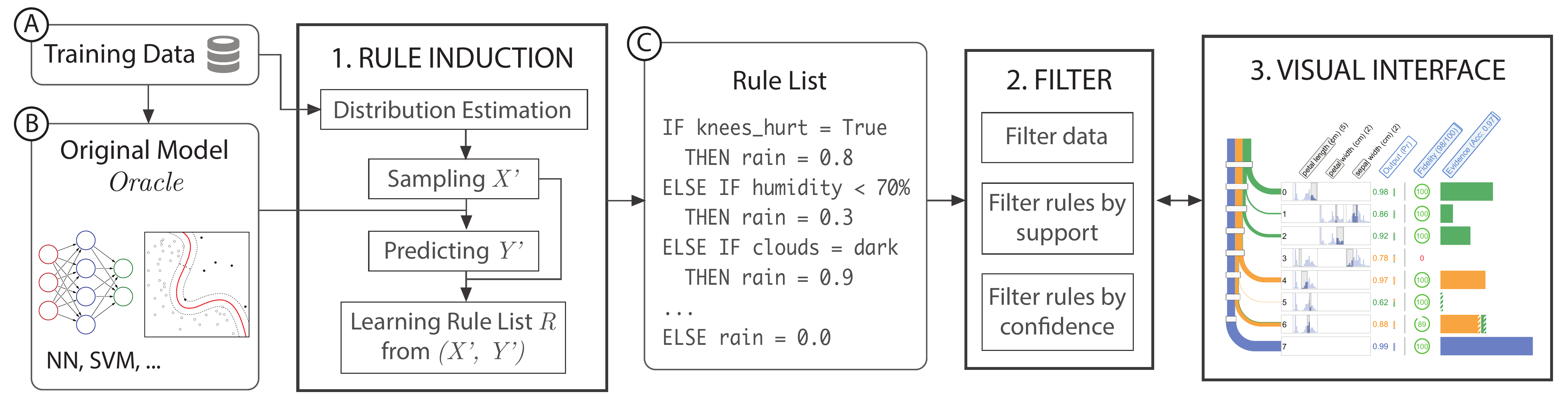}
 \vspace{-1em}
 \caption{The pipeline for creating a rule-based explanation interface. The rule induction step (1) takes (A) the training data and (B) the model to be explained as input, and produces (C) a rule list that approximates the original model. Then the rule list is filtered (2) according to user-specified thresholds of support and confidence. The rule list is visualized as RuleMatrix (3) to help users navigate and analyze the rules. }
 \vspace{-1em}
 \label{fig:pipeline}
\end{figure*}

\section{A Rule-based Explanation Pipeline}

In this section, we introduce our pipeline for creating a rule-based visual interface that helps domain experts understand, validate, and inspect the behavior of a machine learning model. 

\subsection{Goals and Target Users}
In \revise{visualization research, most existing work} for interpreting machine learning models focuses on helping model developers understand, diagnose and refine models. 
In this paper, we target our method at a large number of potential but neglected users -- the experts in various domains that are impacted by the emerging machine learning techniques (\eg, health care, finance, security, and policymakers). 
With the increasing adoption of machine learning in these domains, however, experts may only have little knowledge \revision{of} machine learning algorithms but would like to or are required to use \revision{them} to assist in their decision making.

The primary goal of these potential users, unlike model developers, is to fully understand how a model behaves so that they can better use it and work with it. Before they can fully adopt a model, adequate trust about how the model generally behaves need to be established. Once a model is learned and deployed, they would still need to verify its predictions in case of failure. Specifically, our goal is to help the domain experts answer the following questions:

\begin{enumerate}[label=\textbf{Q\arabic*},leftmargin=0.23in]
\setlength\itemsep{0em}
\item \label{question:knowledge}
\textbf{What knowledge has the model learned?} A trained machine learning model can be seen as an extracted representation of knowledge from the data. 
We propose to present a unified and understandable form of learned knowledge for any given model as rules (\ie, IF-THEN statements). Here each piece of knowledge consists of two parts: the antecedent (IF) and the consequent (THEN). In this way, users can focus on understanding the learned knowledge itself without extra burden of dealing with different representations.

\item \label{question:certain}
\textbf{How certain is the model for each piece of knowledge?} There are two types of certainty that we should consider: the \textit{confidence} (\revise{the probability that a rule is true according to the model}) and the \textit{support} (\revise{the amount of data in support of a rule}). Low confidence means that the rule cannot separate the classes apart, while a low support indicates that there is little evidence for the rule. These are important metrics that help the experts decide whether to accept or reject the learned knowledge.

\item \label{question:verify}
\textbf{What knowledge does the model utilize to make a prediction?} This is the same question as ``Why does the model predict $x$ as $y$''. Unlike the previous two questions, this question is about verifying the model's prediction on a single instance or a subset of instances, instead of understanding the model in general. This is crucial when users prefer to verify the reasons for a model's prediction than to blindly trust it. For example, a doctor would want to understand the reasons of an automatic diagnosis before making a final decision. Domain experts may have knowledge and theories that are originated in years of research and study, which is what current machine learning models fail to utilize.

\item \label{question:fail}
\textbf{When and where is the model likely to fail?} This question arises when a model does not perform well on out-of-sample data. A rule that the model is \revision{confident} about may not be generalizable \revision{in the production}. \revision{Though undesirable, it is not rare that a model gives a highly confident but wrong prediction.} Thus, we need to provide guidance on when and where the model is likely to fail.

\end{enumerate}


\subsection{Rule-based Explanation}
What are explanations of a machine learning model? In existing literature, explanations can take different forms. One widely accepted form of explanation in the machine learning community is the gradients of the input \cite{Simonyan:2014:saliency}, which is often used to analyze deep neural networks. Recently, Ribeiro \etal \cite{Ribeiro:2016:WIT} and Krause \etal \cite{Krause:2017:WVD} defined an explanation of the model's prediction as a set of features that are salient for predicting an input. Explanations can also be produced via analogy, that is, explaining the model's prediction of an instance by providing the predictions of similar instances. These explanations, however, can only be used to explain the model locally for a single instance.


In this paper, we present a new type of explanation that utilizes rules \revise{to explain machine learning models globally} (\autoref{question:knowledge}). A rule-based explanation of a model's prediction $\mathcal{Y}$ of a set of instances $\mathcal{X}$ is a set of IF-THEN decision rules. For example, a model predicts that today it will rain. A human explanation might be: \textit{it will rain because my knees hurt}. The underlying rule format of the explanation would be: \texttt{IF knees\_hurt = True THEN rain = 0.9}. Such explanations with implicit rules occur throughout daily life, and are analogous to the inductive reasoning process that we use every day. 

It should be also noted that there exist different variants of rule-based models. For example, rules can be mutually-exclusive or inclusive (\ie, an instance can fire multiple rules), conjunctive (AND) or disjunctive (OR), and standard or oblique (\ie, contain composite features). 
Though mutually-exclusive rule sets do not require conflict resolution, the complexity of a single rule is usually much larger than an inclusive rule set. In our implementation, we use the representation of an ordered list of \revision{inclusive} rules (\eg, Bayesian Rule Lists \cite{Letham:2015:BRL,Yang:2016:sbrl}). When performing inference, each rule is queried in order and will only fire if all its previous rules are not satisfied. This allows fast queries and bypasses the complex conflicts resolution mechanisms.

\subsection{The Pipeline}

Our pipeline for creating rule-based visual explanations consists of the three steps (\autoref{fig:pipeline}): 1. Rule Induction, 2. Filtering, and 3. Visualization.


\textbf{Rule Induction}. 
Given a model $F$ that we want to explain, the first step is to extract a rule list $R$ that can explain it. There are multiple choices of algorithms as discussed in \autoref{related:model-induction}. In this step we adopt the common pedagogical learning settings. The original model is treated as a teacher, and the student model is trained using the data ``labeled'' by the teacher. That is, we use the predictions of the teacher model as labels instead of the real labels. The algorithm is described in detail in Section 4.

\textbf{Filtering}.
After extracting a rule list approximation of the original model, we will have a semi-understandable explanation. The rule list is understandable in the sense that each rule is human-readable. However, the length of the list can grow too long (\eg, a few hundreds) to be practically understandable. Thus we adopt a step of filtering to obtain a more compact and informative list of rules.

\textbf{Visualization}.
The simplest way to present a list of rules is just to show a list of textual descriptions. However, there are a few drawbacks associated with purely textual representations. First, it is difficult to identify the importance and certainty of each extracted rule (\autoref{question:certain}). Second, it is difficult to perform verification of the model's prediction if the length of the list is long or the number of features is large. This is because the features used in each rule may be different and not aligned \cite{Huysmans:2011:empirical}, which results in a waste of time in aligning features in input and features used in a rule. 

As a solution, we develop RuleMatrix, a matrix-based representation of rules, to help users understand, explore and validate the knowledge learned by the original model. The details of the filtering and visual interface are discussed in Section 5.

\section{Rule Induction}\label{sec:rule-induction}

In this section, we present the algorithm for extracting rule lists from trained classifiers. The algorithm takes a trained model and a training set $\mathcal{X}$ as input, and produces a rule list that approximates the classifier. 

\subsection{The Algorithm}

We view the task of extracting a rule list as a problem of model induction. Given a classifier $F$, the target of the algorithm is a rule list $R$ that approximates model $F$. We define the \textit{fidelity} of the approximate rule list $R$ as its accuracy with the true labels as the output of $F$:

\begin{equation}
fidelity(R)_{\mathcal{X}} = \frac{1}{|\mathcal{X}|} \sum_{\bm{x} \in \mathcal{X}} [F(\bm{x}) = R(\bm{x})],
\label{eq:fidelity}
\end{equation}
where $[F(\bm{x}) = R(\bm{x})]$ evaluates to 1 if $F(\bm{x}) = R(\bm{x})$ and 0 otherwise. The task can be also viewed as an optimization problem, where we are maximizing the fidelity of the rule list.
Unlike common machine learning problems, we have access to the original model $F$, which can be used as an omniscient \textit{oracle} that we can ask for the labels of new data.  Our algorithm highlights the use of the oracle.

The algorithm contains four steps (\autoref{algo:induction}). First, we model the distribution of the provided training data $\mathcal{X}$. We use a joint distribution estimation that can handle both discrete and continuous features simultaneously. Second, we sample a number of data $\mathcal{X}_{sample}$ from the joint distribution. The number of samples is a customizable parameter and can be larger than the amount of original training data. Third, the original model $F$ is used to label the sampled $\mathcal{X}_{sample}$. In the final step, we use the sampled data $\mathcal{X}_{sample}$ and the labels $\mathcal{Y}_{sample}$ to train a rule list. There are a few choices \cite{Marchand:2005:LDL, Fawcett:2008:PRIE, Yang:2016:sbrl} for the training algorithm.

\setlength{\intextsep}{0.7\baselineskip}

\begin{algorithm}[h]
\SetKwInput{Input}{Input}
\SetKwInput{Parameters}{Parameters}
\SetKwInput{Output}{Output}
\Input{model $F$, training data $\mathcal{X}$, rule learning algorithm \textsc{Train}} \Parameters{parameter $n_{sample}$, feature set $\mathcal{S}$}
\Output{A rule list $R$ that approximates $F$}
\Indp
$M \gets$ \textsc{EstimateDistribution}($\mathcal{X}$, $\mathcal{S}$)\;
Draw samples $\mathcal{X}_{sample} \gets$ \textsc{Sample}($M$, $n_{samples}$)\;
Get the labels of $\mathcal{X}_{sample}$ using: $\mathcal{Y}_{sample} \gets F(\mathcal{X}_{sample})$\;
Rule list $R \gets$ \textsc{Train}($\mathcal{X}_{sample}$, $\mathcal{Y}_{sample}$)\;
\Return{R}\;

 \caption{Rule Induction}
 \label{algo:induction}
\end{algorithm}
\setlength{\intextsep}{\baselineskip}

The distribution estimation and sampling steps are inspired by TrePan \cite{Craven:1995:Trepan}, a tree induction algorithm that recursively extracts a decision tree from a neural network. The sampling is mainly needed for two reasons. First, since the goal is to extract a rule list that approximates the given model, the rule list should also be able to approximate the model's behavior on input that has not been seen before. The sampling helps generate unforeseen data. Second, when the training data is limited, the sampling step creates sufficient training samples, which helps achieve a good fidelity for the extracted rule list. Next, we introduce the details of the algorithm.

\setlength{\intextsep}{0.7\baselineskip}
\begin{algorithm}[h]
\SetAlgoNoLine
\SetKwInput{Input}{Input}
\SetKwInput{Parameters}{Parameters}
\SetKwInput{Output}{Output}
\Input{training data $\mathcal{X}$, feature set $\mathcal{S}$}
\Output{The distribution estimation $M$}
\Indp
Divide the features $\mathcal{S}$ into discrete features $\mathcal{S}_{disc}$ and continuous features $\mathcal{S}_{con}$\;
Partition $\mathcal{X}$ to $\mathcal{X}_{disc}$ and $\mathcal{X}_{con}$ according to $\mathcal{S}_{disc}$ and $\mathcal{S}_{con}$\;
\tcc{Estimate the categorical distribution $p$}
Initialize a counter $Counter: \bm{x}_{disc} \mapsto 0$\;
\For{$\mathbf{x}_{disc}^{(i)}$ in $\mathcal{X}_{disc}$} {
    $Counter[\mathbf{x}_{disc}^{(i)}] \gets Counter[\mathbf{x}_{disc}^{(i)}] + 1$
}
\For{$\mathbf{x}_{disc}^{(i)}$ in $Counter$}{
	$p_{\mathbf{x}_{disc}^{(i)}} \gets Counter[\mathbf{x}_{disc}^{(i)}] / |\mathcal{X}|$\;
}
\tcc{Estimate conditional density $f$}
Estimate the bandwidth matrix $\mathbf{H}$ from $\mathcal{X}_{con}$\;
\For{$\mathbf{x}_{disc}^{(i)}$ in $Counter$}{
	$f_{\mathbf{x}_{disc}^{(i)}} \gets $ \textsc{DensityEstimation}($\mathcal{X}_{con}$, $\mathbf{H}$)\;
}
\Return{$M = (p, f)$}\;

\caption{Estimate Distribution}
 \label{algo:estimate}
\end{algorithm}
\setlength{\intextsep}{\baselineskip}

\paragraph{Distribution Estimation.}
The first step is to build a model $M$ that estimates the distribution of the training set $\mathcal{X} = \{\mathbf{x}^{(i)}\}_{i=1}^N$ with $N$ instances, where each $\mathbf{x}^{(i)} \in \mathbb{R}^k$ is a $k$ dimensional vector. Without losing generality, we assume the $k$ features are mixed with $d$ discrete features $\bm{x}_{disc} = (x_1, ..., x_d)$ and $(k-d)$ continuous features $\bm{x}_{con} = (x_{d+1}, ..., x_k)$. Using Bayes' Theorem, we can write the joint distribution of the mixed discrete and continuous random variables as:
\begin{equation}
\begin{aligned}
f(\bm{x}) =& f(\bm{x}_{disc}, \bm{x}_{con}) \\ 
=& Pr(\bm{x}_{disc}) f(\bm{x}_{con} \mid \bm{x}_{disc}).
\end{aligned}
\end{equation}
The first term is the probability mass function of the discrete random variables, and the second term is the conditional density function of the continuous random variables given the values of the discrete variables. Next we discuss the two terms separately.

We assume that the discrete features $\bm{x}_{disc}$ follow categorical distributions. The probability of each combination of $\bm{x}_{disc}$ can be estimated using its frequency in the training data (\autoref{algo:estimate}, lines 3-9):
\begin{equation}
Pr(\bm{x}_{disc}=\mathbf{x}_{disc}) = \hat{p}_{\mathbf{x}_{disc}} = \frac{\sum_{i=1}^N [ \mathbf{x}_{disc}^{(i)} = \mathbf{x}_{disc}]}{N},
\label{eq:categorical}
\end{equation}
where $[\mathbf{x}_{disc}^{(i)} = \mathbf{x}_{disc}]$ evaluates to 1 if $\mathbf{x}_{disc}^{(i)} = \mathbf{x}_{disc}$, and 0 otherwise. 


We use multivariate density estimation with Gaussian kernel to model continuous features $\mathbf{x}_{con}$ (\autoref{algo:estimate}, line 10-13). Since we are interested in the conditional distribution, we can write the conditional density estimation as:
\begin{equation}
\begin{aligned}
f(&\bm{x}_{con} \mid \bm{x}_{disc}) \\
=&\frac{1}{|S|}\sum_{\mathbf{x} \in S} 
\frac{
\exp\{-\frac{1}{2}(\bm{x}_{con}- \mathbf{x}_{con})^T \mathbf{H}^{-1} (\bm{x}_{con} - \mathbf{x}_{con})\}
}{
(2\pi)^{\frac{c}{2}} |\mathbf{H}|^{\frac{1}{2}}
},
\label{eq:kde}
\end{aligned}
\end{equation}
where $S = \{ \mathbf{x} \mid \mathbf{x}_{disc}=\bm{x}_{disc}, \mathbf{x}\in\mathcal{X} \}$ is a subset of training data that has the same discrete values as $\bm{x}_{disc}$, and $c = (k-d)$ is the number of the continuous features.
Here $\mathbf{H}$ is the bandwidth matrix, and also the covariance matrix for the kernel function. 
The problem left is how to choose the bandwidth matrix $\mathbf{H}$. There are a few methods for estimating the optimal choice of $\mathbf{H}$, such as smoothed cross validation and plug-in. For simplicity, we adopt Silverman's rule-of-thumb \cite{silverman:1986:density}:
\begin{equation}
\begin{aligned}
\sqrt{\mathbf{H}_{ii}} &= (\frac{c+2}{4} n)^{-\frac{1}{c+4}} \sigma_i \\
\mathbf{H}_{ij} &= 0,  \quad i \neq j,
\end{aligned}
\end{equation}
where $\sigma_i$ is the standard deviation of feature $i$.

Once we have built a model of the distribution, $M$, we can easily create $\mathcal{X}_{sample}$. The question left is how to choose a proper number of samples, which will be discussed in \autoref{sec:algo-experiments}.



\paragraph{Rule List.}
In the last step, a training algorithm \textsc{Train} is needed to learn a rule list from $(\mathcal{X}_{sample},\mathcal{Y}_{sample})$. There exist various algorithms that can construct a list of rules from training data \cite{Marchand:2005:LDL,Fawcett:2008:PRIE,wang:2015:frl,Yang:2016:sbrl}. Both of the algorithms proposed by Marchand and Sokolova \cite{Marchand:2005:LDL} and Fawcett \cite{Fawcett:2008:PRIE} follow a greedy construction mechanism and do not offer a good performance. In the implementation, we adopt the Scalable Bayesian Rule List (SBRL) algorithm proposed by Yang \etal \cite{Yang:2016:sbrl}. 
This algorithm models the rule list using a Bayesian framework and allows users to specify priors related to the length of the list and the complexity of each rule. This is useful for our task, since we can have controls on the complexity of the extracted rule list. This algorithm also has the advantage that it can be more naturally extended to support multi-class classification (\ie, by switching the output distribution from binomial to multinomial), which supports a more generalizable solution. Readers can refer to the paper by Yang \etal \cite{Yang:2016:sbrl} for more details.

Note that the algorithm requires a preprocessing step to discretize the input and pre-mine a candidate rule sets for the algorithm to choose from. In our implementation, we use the minimum description length (MDL) discretization \cite{Rissanen:1978:MDL} to discretize continuous features, and use the FP-Growth item set mining algorithm \cite{Han:2000:FPGrowth} to get the candidate rule sets. Other discretization and rule mining methods can also be used.

\subsection{Experiments}\label{sec:algo-experiments}

To study the effect of sample size and evaluate the performance of the proposed rule induction algorithm, we test our induction algorithm on several publicly available datasets from the UCI Machine Learning Repository \cite{uci:2017} and a few popular models that are commonly regarded as hard to interpret. 

\begin{figure}[bt]
 \centering 
 \includegraphics[width=1.0\columnwidth]{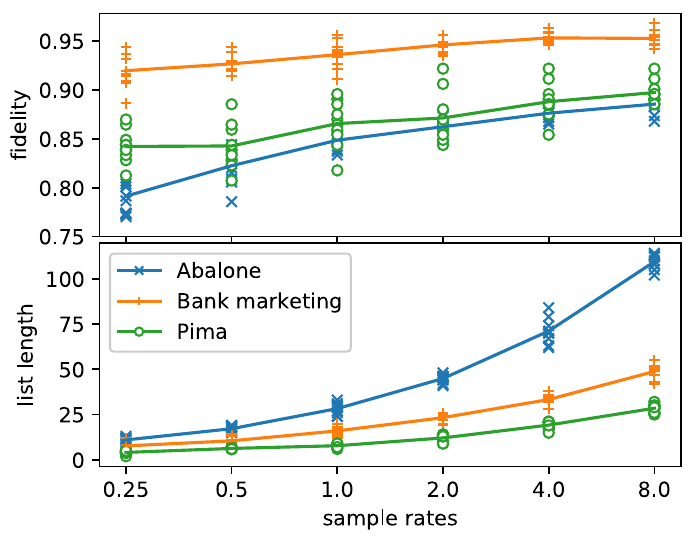}
  \vspace{-1.6em}
 \caption{The performance of the algorithm under different sampling rates. \revision{The x-axis shows the logarithms of the sampling rates. The \textcolor{blue}{blue}, \textcolor{orange}{orange}, and \textcolor{green}{green} lines show the average fidelities and average lengths of the extracted rule lists on the Abalone, Bank Marketing and Pima datasets for 10 runs.} }
  \vspace{-0.5em}
 \label{fig:sample-rate}
\end{figure}

\textbf{Sampling Rate}. First, we study the effect of \textit{sampling rate} (\ie, number of samples / number of training data) using three datasets, Abalone, Bank Marketing and Pima Indian Diabetes (Pima). Abalone contains the physical measurements of 4177 abalones originally labeled with their rings (representing their ages). Since our current implementation only supports classification, we replace the number of rings with four simplified and balanced labels, \ie, $rings < 9$, $9 \leq rings < 12$, $12 \leq rings < 15$, and $15 < rings$, with 1407, 1810, 596, and 364 instances respectively. Bank Marketing and Pima are binary classifications. All three datasets are randomly partitioned into a 75\% training set and a 25\% test set. We train a neural network with four hidden layers and 50 neurons per layer on the training set. Then we test the algorithm on the neural network with six sampling rates growing exponentially: 0.25, 0.5, 1.0, 2.0, 4.0, and 8.0. We run each setting 10 times and compute the fidelity on the test set. 

As shown in \revision{\autoref{fig:sample-rate}}, with all three datasets, the fidelity of extracted rule lists generally increases as the sampling rate grows. However, the complexity of the rule lists also increases dramatically (which is also a reason for an additional visual interface). Here there is a trade-off between the fidelity and interpretability of the extracted rule list. Considering that interpretability is our major goal, we adopt the following strategy for choosing sampling rate: start from a small sampling rate (1.0), and gradually increase the sampling rate until we get a good fidelity or the length of the rule list exceeds an acceptable threshold (\eg, 60).

\begin{table}[tb]
  \caption{The fidelities of the rule list generated by the algorithm from a neural network and an SVM. The table reports the mean and standard deviation (with parenthesis) in percentage of the fidelity of 10 runs for each setting.}
  \label{tab:algo-experiment}
  \scriptsize%
	\centering%
  \begin{tabu}{%
	r%
	*{4}{c}%
	}
  \toprule
  Dataset & NN-1 & NN-2 & NN-4 & SVM \\
  \midrule
	Breast Cancer & 95.5 (1.4) & 94.5 (1.5) & 95.0 (2.0) & 95.9 (1.4) \\
    Wine & 93.1 (2.3) & 94.0 (2.4) & 94.0 (3.7) & 91.3 (3.5) \\
    Iris & 96.3 (1.7) & 97.9 (2.6) & 94.7 (3.1) & 97.4 (2.0) \\
    Pima & 89.6 (2.0) & 89.9 (1.2) & 89.5 (1.7) & 91.8 (1.5) \\
    Abalone & 88.5 (0.9) & 88.6 (0.7) & 86.8 (0.5) & 90.1 (0.8) \\
    Bank Marketing & 96.4 (0.8) & 92.1 (1.0) & 89.1 (1.3) & 97.0 (0.7) \\
    Adult & 95.0 (0.2) & 94.8 (0.4) & 93.2 (0.3) & 96.7 (0.3) \\
  \bottomrule
  \end{tabu}%
   \vspace{-0.5em}
\end{table}

\textbf{Fidelity}. To verify that the proposed rule induction algorithm is able to produce a good approximation of a given model, we benchmark the algorithm on a set of datasets with two different classifiers, neural networks and support vector machines.
The datasets we use include: Breast Cancer Wisconsin (Diagnostics), Iris, Wine, Abalone (four-class classification), Bank Marketing, Pima Indian Diabetes and Adult. 

\begin{figure*}[t]
 \centering 
 \includegraphics[width=\textwidth]{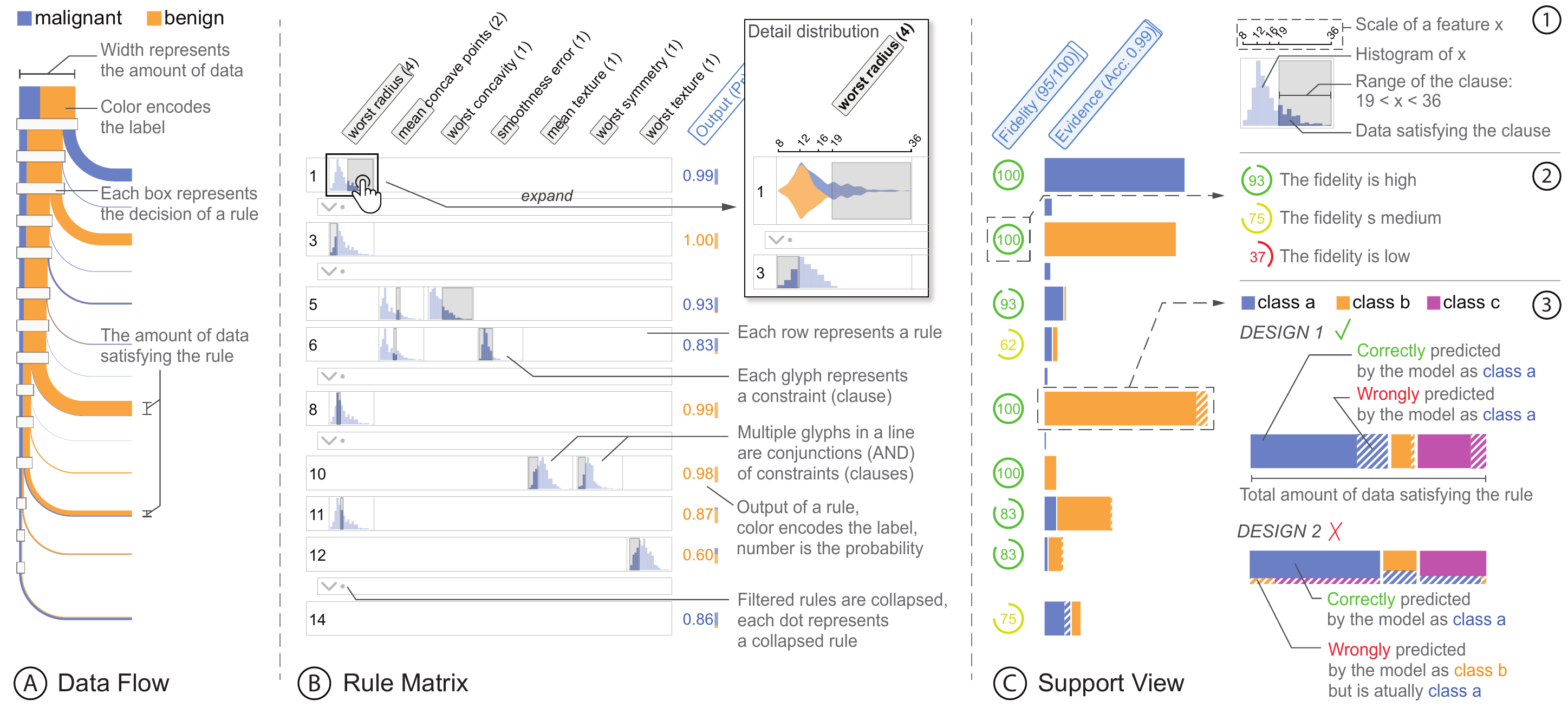}
 \vspace{-1.2em}
 \caption{The visualization design. A: The data flow visualizes the data that satisfies a rule as a flow into the rule, providing an overall sense of the order of the rules. B: The rule matrix presents each rule as a row and each feature as a column. The clauses are visualized as glyphs in the corresponding cells. Users can click to expand a glyph to see the details of the distribution and the interval of the clause.  C: The support view shows the fidelity of the rule for the provided data, and the evidence of the model's predictions and errors under a certain rule. \circletext{1} - \circletext{3}: The designs of the glyph, the fidelity visualization, and the evidence visualization.}
 \vspace{-0.6em}
 \label{fig:visual-design}
\end{figure*}

We test the algorithm on neural networks with one, two, and four hidden layers, and support vector machines with nonlinear Radial Basis Function (RBF) kernel. We use the implementation of these models in the scikit-learn package \cite{scikit-learn}. We use a sampling rate of 2.0 for the Adult dataset, and a sampling rate of 4.0 for the rest. As shown in \autoref{tab:algo-experiment}, the rule induction algorithm can generate rule lists that approximate a model with acceptable fidelity on the selected datasets. The fidelity is over 90\% on most datasets except for Pima and Abalone. 

\textbf{Speed}. The time for creating a list of 40 rules from 7,000 samples with 20 features can take up to 3 minutes on a PC (the time varies under different parameters). The estimation and sampling step take less than one second, and the major bottleneck lies in the FP-Growth (less than 10 seconds) and SBRL (more than 2 minutes) algorithms. We restrict the discussion of this issue in this paper due to page limits. The material necessary for reproduce the results is available at \url{http://rulematrix.github.io}.




\section{RuleMatrix: The Visual Interface} \label{sec:visual}
This section presents the design and implementation of the visual interface for helping users understand, navigate and inspect the learned knowledge of classifiers. As shown in \autoref{fig:teaser}, the interface contains a control panel (A), a main visualization (B), a data filter panel (C) and a data table (D). In this section, we mainly present the main visualization, RuleMatrix, and the interactions supported by the other views.




\subsection{Visualization Design}
RuleMatrix (\autoref{fig:visual-design}) consists of three visual components: the \textit{rule matrix}, the \textit{data flow}, and the \textit{support view}. The rule matrix visualizes the content of a rule list in a matrix-based design. The data flow shows how data flows through the list using a Sankey diagram. The support view supports the understanding and analysis of the original model that we aim to explain.

\subsubsection{Rule Matrix}
The major visual component of the interface is the matrix-based visualization of rules. A decision rule is a logical statement consisting of two parts: the \textit{antecedent} (IF) and the \textit{consequent} (THEN). Here we restrict the antecedent to be a conjunction (AND) of clauses, where each clause is a condition on an input feature (\eg, $3 < x_1$ AND $x_2 < 4$). This restriction eases users' cognitive burden of discriminating different logical operations. The output of each rule is a probability distribution over possible classes, representing the probability of an instance satisfying the antecedent belongs to each class. The simplest way to present a rule is to write it down as a logical expression, which is ubiquitous in programing languages. However, we found textual representations difficult to navigate when the length of the list is too large. The problem with textual representations is that the input features are not presented in the same order in each rule. Thus, it is difficult for users to search a rule with certain condition or compare the conditions used in different rules. This problem has also been identified by Huysmans \etal \cite{Huysmans:2011:empirical}, 

To address this issue and help users understand and navigate the rule list (\autoref{question:knowledge}), we present the rules in a matrix format. As shown in \autoref{fig:visual-design}B, each row in the matrix represents the antecedent of a decision rule, and each column represents an input feature. If the antecedent of a decision rule $i$ contains a clause using feature $x_j$, then a compact representation (\autoref{fig:visual-design}-\circletext{1}) of the clause is shown in the corresponding cell $(i,j)$. In this layout, the order of the features is fixed, which helps users visually search and compare rules by features. 
The length of the bar underneath a feature name encodes the frequency with which the feature occurs in the decision rules. The features are also sorted according to their importance scores, which is computed by the number of instances that a feature has been used to discriminate. The advantage of the matrix representation is that it allows users to verify and compare different rules quickly. This also allows easier verification and evaluation of the model's predictions (\autoref{question:verify}).

\textbf{Visualizing Conditions.} In the antecedent of rule $i$, a clause that uses feature $j$ (\eg, $0 \leq x_j < 3$) is visualized as a gray and translucent box in cell $(i, j)$, where the covered range represents the interval in the clause (\ie, $[0, 3)$). In each cell $(i,j)$, a compact view of the data distribution of feature $j$ is also presented (inspired by the idea of \textit{sparklines} \cite{Tufte:2006:BE}). For continuous features, the distributions are visualized as histograms. For discrete features, bar charts are used. The part of data that satisfies the clause is also highlighted with a higher opacity. This combination of the compact view of data distribution and the range constraint helps users quickly grasp the properties of different clauses in a rule (\autoref{question:knowledge}), \ie, the tightness or width of the interval and the number of instances that satisfy the clause.

\textbf{Visualizing Outputs.} As discussed above, the output of a rule is a probability distribution. At the end of each row, we present the output of the rule as a colored number, with color representing the output label of the rule, and the number showing the probability of the label. A vertically stacked bar is positioned next to the number to show the detailed probability of each label. Using this design, users are able to quickly identify the output label of the rule by the color, and learn the actual probability of the label from the number.

\subsubsection{Data Flow}
To provide users with an overall sense of how the input data is classified by different rules, a waterfall-like Sankey diagram (\autoref{fig:visual-design}A) is presented to the left of the rule matrix. The main vertical flow represents the data that remains unclassified. Each time the main flow ``encounters'' a rule (represented by a horizontal bar), a horizontal flow representing the data satisfying the rule forks from the main vertical flow. The widths of the flows represent the quantities of the data. The colors encode the labels of the data. That is, if a flow contains data with multiple labels, the flow is divided into multiple parallel sub-flows, whose widths are proportional to the quantities of different labels. The data flow helps the user maintain a proper mental model of the ordered decision rule list. The rules are ordered, and the success of a rule has the implication that previous rules are not satisfied. The user can identify the amount of data satisfying a rule \revision{through the width of the flow, which helps the user decide to trust or reject the rule (\autoref{question:certain}). The design of the data flow is inspired by the node-link layout used in BaobabView \cite{elzen:2011:baobabview}.}



\subsubsection{Support View}
The support view is designed to support the understanding and analysis of the performance of the original model. Note that there are two types of errors that we are interested in: the error between the rule and the model (fidelity), and the error between the model and the real data (accuracy). When the error between a rule and the model is high, users should be notified that the rule may not be a well-extracted ``knowledge''. When the error between the original model and the real data is high, the users should be notified that the model's prediction should not be fully trusted (\autoref{question:fail}). In the support view, we provide for each rule a set of two performance visualizations (\autoref{fig:visual-design}C), fidelity and evidence to help users analyze these two types of errors. 

\textbf{Fidelity}. We use a simple glyph that contains a number (0 to 100) to present the \textit{fidelity} (\autoref{eq:fidelity}) of the rule on the subset of data satisfying the rule. The value of fidelity represents how accurately the rule represents the original model on this subset. The higher the fidelity, the more reliable the rule is in representing the original model. The number is circled by an arc, whose angle also encodes the number. As shown in \autoref{fig:visual-design}-\circletext{2}, the glyph can be colored green (high), yellow (medium), red (low) according to the level of fidelity. In the current implementation, the fidelity levels are set to above 80\% (high), 50\% (medium) to 80\%, and below 50\% (low), respectively. 

\textbf{Evidence}. The second performance visualization shows the \textit{evidence} of the original model on the real data (users can switch between training or test set). To support comprehensive analysis of the error distribution, we adopt a compact and simplified variant of Squares \cite{ren:2016:squares}. As shown in Design 1 in \autoref{fig:visual-design}-\circletext{3}, we use horizontally stacked boxes to present the predictions of the model. The color encodes the predicted class by the original model. The width of a box encodes the amount of data with a certain type of prediction. We use striped boxes to represent erroneous predictions. That is, a blue striped box (\bluestripe{}) represents data that is wrongly classified as class blue and has real labels different from class blue. During the development of this interface, we have experimented with an alternative design which had the same color coding, as shown in Design 2 in \autoref{fig:visual-design}-\circletext{3}. In this alternative design, the data is divided into horizontally stacked boxes according to the true labels. Then we partition each box vertically into two parts: the upper one representing correct predictions and the lower one representing the wrong predictions (striped boxes). The lower part is further partitioned into multiple parts according to the predicted labels. However, during our informal pilot studies with two graduate students, the Design 2 was found to be ``confusing'' and ``distracting''. Though Design 1 fails to present the real labels of the wrong predictions, it is more concise and can be directly used to answer whether a model is likely to fail (\autoref{question:fail}).

The advantage of the compact performance visualization is that it presents an intuitive error visualization within a small space. We can easily identify the amount of instances classified as a label or quantify the mistakes by searching for the boxes with the corresponding coding.

\subsection{Interactions}
RuleMatrix supports three types of interactions: filtering the rules, which is used to reduce cognitive burden by reducing the number of rules to show; filtering the data, which is used to explore the relation between the data and the rules; and details on demand.

\subsubsection{Filtering the Rules}
The filtering of rules helps relieve the scalability issue and reduce the cognitive load when the extracted rule list is too long. This occurs when we have a complex model (\eg, a neural net with multiple layers, or an SVM with nonlinear kernel), or a complex data set. In order to learn a rule list that well approximates the model, the complexity of the rule list inevitably grows. In our implementation, we provide two types of filters: \textit{filter by support} and \textit{filter by confidence}. The former filters the rules that have little support, which are seldom fired and are not salient. The latter filters the rules that have low confidence, which are not significant in discriminating different classes. In our implementation, filtered rules are grouped into collapsed ``rules'' so that users can keep track of them. Users can also expand the collapsed rules to see them in full details. By adjusting rule filters, users are allowed to explore a list of over 100 rules with no major cognitive burden.

\subsubsection{Filtering the Data}
The data filtering function is needed to support two scenarios. First, data filtering allows users to apply the \textit{divide and conquer} strategy to understand the model's behavior, \ie, only focus on the model's behavior on the data one is interested in. Second, by filtering, users can identify the data entries in the data table (\autoref{fig:teaser}D) that support specific rules. This boosts users' trust in both the system and the model. During our experiments, we found that data filters can greatly reduce the number of rules shown when combined with rule filters.

\subsubsection{Details on Demand}
To provide a clean and concise interface, we hide the details that users can view on demand. Users can request details in two ways: interacting with the RuleMatrix directly or modifying the settings in the control panel. 
In the RuleMatrix, users can check the actual text description of a clause by hovering on the corresponding cell. To view the details about the data distribution, users can click on a cell, which expand the cell and show a stream plot (continuous feature) or a stacked bar charts (categorical feature) of the distribution (\autoref{fig:visual-design}B). \revision{The choice of stream plot for continuous features is due to its ability in preventing color discontinuities \cite{elzen:2011:baobabview}.}  A vertical ruler that follows the mouse is displayed to help align and compare the intervals of the clauses using the same feature across multiple rules. Users can see the actual amount of data by hovering over the evidence bars or certain parts of the data flow. Users can view the conditional distribution or hide the striped error boxes by modifying the settings in the control panel. Here the conditional distribution of feature $x_j$ at rule $i$ denotes the distribution given that all previous rules are not satisfied, that is, the distribution of the data that is left unclassified until rule $i$.

The rule filtering functions are provided in the control panel (\autoref{fig:teaser}A), and the data filtering functions are provided in the data filter (\autoref{fig:teaser}C). Users are also allowed to customize an input and request the system to present the prediction of the original model and highlight the satisfied rule.

\section{Evaluation}
We present a usage scenario, a use case, and a user study to demonstrate how our method effectively helps users understand the behavior of a classifier.

\begin{figure}[hbt]
 \centering 
 \includegraphics[width=1.0\columnwidth]{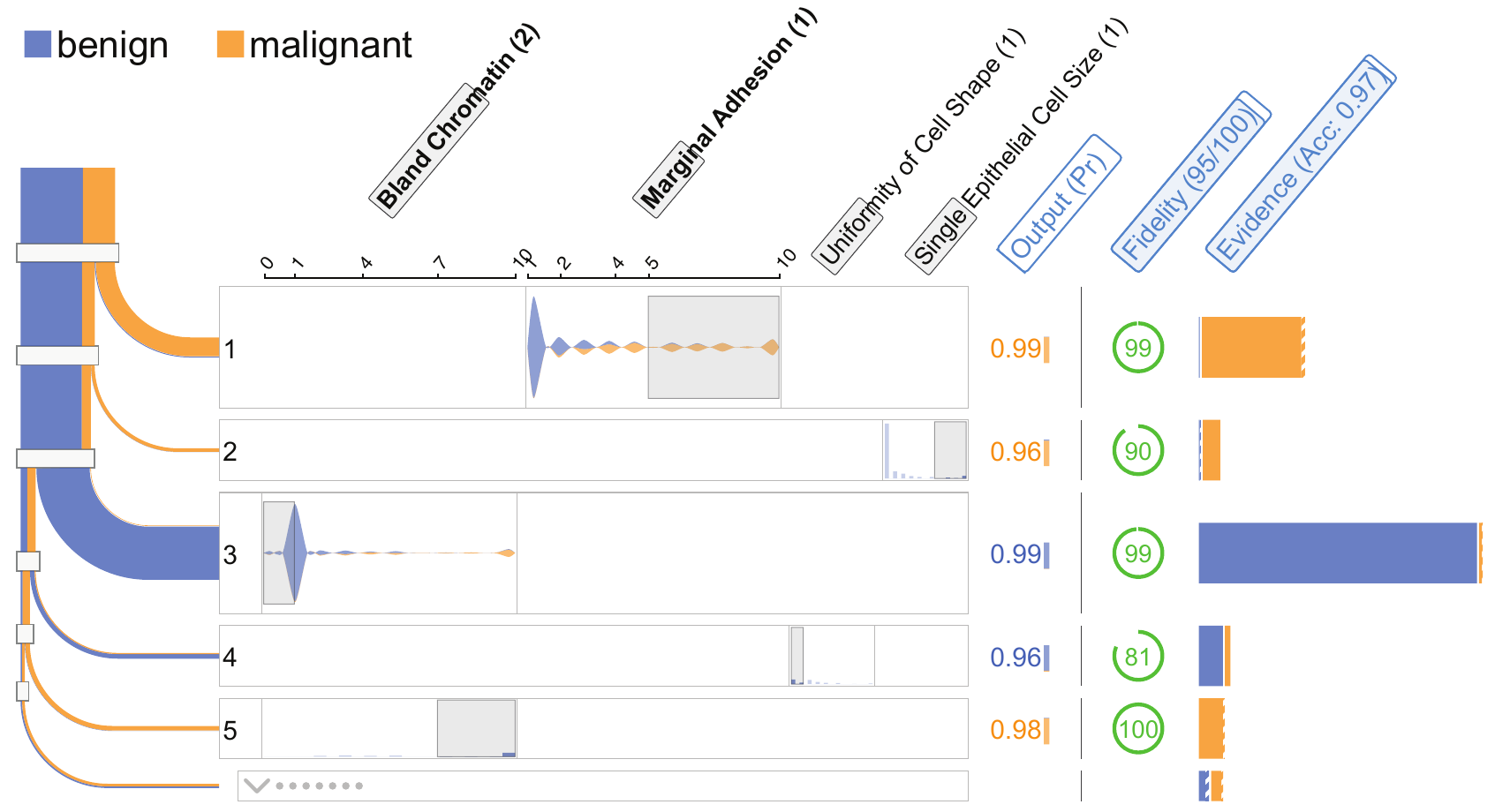}
 \caption{Using the RuleMatrix to understand a neural network trained on the Breast Cancer Wisconsin (Original) dataset.}
 \vspace{-0.5em}
 \label{fig:case-cancer}
\end{figure}

\begin{figure*}[t]
 \centering 
 \includegraphics[width=2\columnwidth]{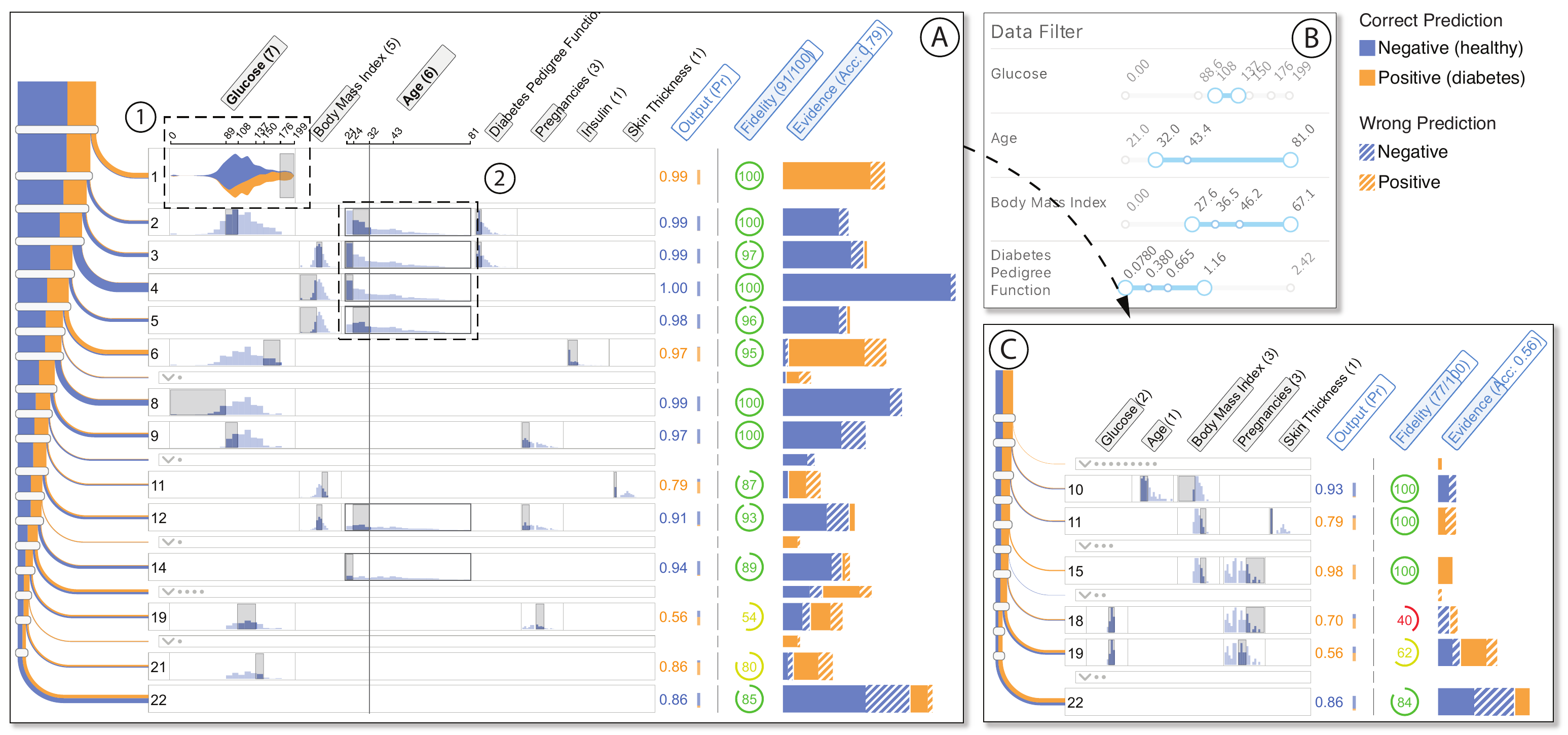}
 \caption{The use case of understanding a neural network trained on Pima Indian Diabetes dataset. A: The initial visualization of the list of 22 extracted rules, with an overall fidelity of 91\%. The neural network has an accuracy of 79\% on the training data. B: The applied data filter. The ranges of the features are highlighted with light blue. C: The visualization of the rule list with the filtered data. The accuracy of the original model drops to only 56\%.}
 \vspace{-0.5em}
 \label{fig:case-pima}
\end{figure*}

\subsection{Usage Scenario: Understanding a Cancer Classifier}
We first present a hypothetical scenario to show how RuleMatrix helps people understand the knowledge learned by a machine learning model. 

Mary, a medical student is learning about breast cancer and is interested in identifying cancer cells by the features measured from biopsy specimens. She is also eager to know whether the popular machine learning algorithms can learn to classify cancer cells accurately. She downloads a pre-trained neural network and the Breast Cancer Wisconsin dataset from the Internet. \revision{The dataset contains cytological characteristics of 699 breast fine-needle aspirates. Each of the cytological characteristics are graded from 1 to 10 (lower is closer to begin) at the time of sample collection. The accuracies of the model on training and test set are 97.3\% and 97.1\% respectively.  She want to know what knowledge the model has learned (\autoref{question:knowledge}).} 

\textbf{Understanding the rules}. Mary uses our pipeline and extracts a list of 12 rules from the neural network. The visualization is presented to Mary. She quickly goes through the list and notices that rule 6 to rule 12 have little support from the training data (\autoref{question:certain}). Then she adjust the minimum evidence in the rule filter (\autoref{fig:teaser}A) to 0.014 to collapse the last 7 rules (\autoref{fig:case-cancer}). She then finds that the first rule outputs \textcolor{visorange}{malignant} with a high probability (0.99) and a high fidelity (0.99). She looks into the rule matrix and learns that if the marginal adhesion score is larger than 5, the model will very likely predict \revision{malignancy}. This aligns with her knowledge that the loss of adhesion is a strong sign of cancer cells. Then she checks rule 3, which has the largest support from the dataset. The rule shows that if the bland chromatin \revision{(the texture of nucleus)} is smaller or equal than 1, the cell should be \textcolor{visblue}{benign}. She finds this rule interesting since it indicates that one can quickly identify benign cells in the examination by checking if the nucleus \revision{is coarse}.


\subsection{Use Case: Improving Diabetes Classification}
In this use case, we \revision{the Pima Indian Diabetes Dataset (PIDD) \cite{smith1988pima} to} demonstrate how RuleMatrix can lead to performance improvements. The dataset contains diagnostic measurements of 768 female patients aged from 21 to 81, of Pima Indian heritage. The task is to classify negative patients (healthy) and positive patients (has diabetes). Each data instance contains eight features: the number of previous pregnancies, plasma glucose, blood pressure, skin thickness, insulin, body mass index (BMI), and diabetes pedigree function (DPF). DPF is a function measuring a patient's probability of getting diabetes based on the history of the patient's ancestors. The dataset is randomly partitioned into 75\% training set and 25\% test set. 
The distribution of the labels in the training set and test set are 366 negatives / 210 positives and 134 negatives / 58 positives respectively.

In the beginning, we trained a neural network of 2 layers with 20 neurons in each layer. The l-2 normalization factor was determined as 1.0 via 3-fold cross-validations. We ran the training 10 times and received an average accuracy of 72.4\% on the test data. The best neural network had an accuracy of 74.0\% on the test set. We ran the proposed rule-based explanation pipeline and extracted a list of 22 decision rules from a trained network. The rule list is visualized with the training data and a rule filter of minimum evidence of 0.02 (\autoref{fig:case-pima}A). From the header ``evidence'', we can see that the neural network achieves an overall accuracy of 79\% on the training set. 

\textbf{Understanding the Rules} (\autoref{question:knowledge}, \autoref{question:certain}). Then we navigated the extracted rules using the RuleMatrix \revision{with the training set}.
We noticed that there was no dominant rules with \revision{large supports}, except for rule 4 and the last default rule, which have relatively longer bars in the ``evidence'' column, indicating a larger support. This reflects that the dataset is in a difficult domain and it is not easy to accurately predict whether one has diabetes or not. Rule 1 (\autoref{fig:case-pima}-\circletext{1}) has only one condition, 176 $<$ plasma glucose, which means that a patient with high plasma glucose is very likely to \textcolor{visorange}{have diabetes}. This agrees with our common knowledge in diabetes. Then we noticed that the outputs of rules 2 to 5 were all \textcolor{visblue}{negative} with probabilities above 0.98. Thanks to the aligned layout of features, we derived an overall sense that the patients younger than 32 (\autoref{fig:case-pima}-\circletext{2}) and a BMI less than 36.5 are not likely to have diabetes. After going through the rest of the list, we concluded that patients with high plasma glucose and high BMI are more likely to have diabetes, and young patients are less likely to have diabetes in general.

\textbf{Understanding the Errors} (\autoref{question:fail}). After navigating the rules, we were mostly interested in the type of patients that have diabetes but are wrongly classified as negative by the neural network. The false negative errors are undesirable in this domain since they may delay the treatment of a real patient and cause higher risks. Based on our findings concluded from rules 2 to 5, we decided to focus on the patients older than 32, that is, those with higher risk. We also filtered the patients with low or high plasma glucose (lower than 108 or higher than 137), because most of them are correctly classified as negative or positive by the model. As a result of the filtering, the accuracy of the model on the remaining data (74 instances) immediately dropped to 62\%. From the resulting rules, we then further filtered patients with a BMI lower than 27, who are unlikely to have diabetes, and the patients with a DPF higher than 1.18, who are very likely to have the disease. After the filtering (\autoref{fig:case-pima}B), the accuracy of the model on the resulting subset of 62 patients dropped to only 56\%. From \autoref{fig:case-pima}C, we found a large portion of blue striped boxes (\bluestripe{}), denoting patients that have diabetes but were wrongly classified as healthy. This validated our suspicion that the patients with no obvious indicators are difficult to classify.

\textbf{Improving the Performance}. Based on the understanding of the error, a simple idea appeared to be worth trying: can we improve the accuracy of the model by oversampling the difficult subset? We experimented by oversampling this subset by half the amount to get 31 new training data, and trained new neural networks with the same hyper-parameters. To determine whether the change led to an actual improvement, we ran the training and sampling 10 times. The mean accuracy of 10 runs reached 75.5\% on the test set, with a standard deviation of 2.1\%. The best model had a performance of 78.6\%, which was significantly better than the original best model (74.0\%). 

\subsection{User Study}
We conducted a quantitative experiment to evaluate the effectiveness of RuleMatrix in helping users understand the behavior of machine learning models. The main goal of the experiment is to investigate whether users can understand the interactive matrix-based representation of rules, and whether users can understand the behavior of a given model via the rule-based explanations. We asked participants to perform relevant tasks to benchmark the effectiveness of the proposed interface, and asked for subjective feedbacks to understand users' preference and directions for improvements.

\textbf{Study Design}. We recruited nine participants, ages 22 to 30. Six were current graduate students majoring in computer science, three had experience in research projects related to machine learning, and none of them had prior experiences in model induction. \revision{The experiments were conducted using a 23'' monitor.} 

The study was organized into three steps. First, each participant was presented with a 15 minutes tutorial and was given 5 minutes to navigate and explore the interface.
Second, participants were asked to perform a list of tasks using RuleMatrix. 
Finally, participants were asked to answer five subjective questions related to the general usability of the interface and suggestions for improvements. 
We used the Iris dataset and an SVM as the to-be-explained model during the tutorial. In the formal study, we used the Pima Indian Diabetes dataset, and used RuleMatrix to explain a neural network with two hidden layers with 20 neurons per layer. The extracted rule list contained 20 rules, each containing a conjunction of 1, 2, or 3 clauses).


\bgroup
\def\arraystretch{1.5}
\begin{table}[tb]
  \caption{The experiment tasks and results. The results are summarized as the number of correct answers / total number of questions.}
  \label{tab:tasks}
  \scriptsize%
	\centering%
  \begin{tabu}{%
	p{0.1cm} p{0.2cm} p{6.2cm} p{0.5cm}
	}
  \toprule
   & Goal & Question & Result \\
  \midrule
	T1 & \autoref{question:knowledge} & Which of the textual descriptions best describe rule $i$? & 16/18 \\
    T2 & \autoref{question:knowledge} & Which of the rules exists in the extract rule lists? & 18/18 \\
    T3 & \autoref{question:certain} & Which of the highlighted rules is most reliable in representing the original model? & 17/18 \\
    T4 & \autoref{question:certain} & Which of the highlighted rules has the largest support? & 17/18 \\
    T5 & \autoref{question:fail} & Under which of the four highlighted rules, the original model is most likely to give wrong predictions? & 17/18 \\
    T6 & \autoref{question:verify} & For a given data (presented in texts), 
    \newline (a) what would the original model most likely to predict? 
    \newline (b) which rule do you utilize to perform the prediction? & \t \newline 18/18 \newline 17/18 \\

  \bottomrule
  \end{tabu}%
\end{table}
\egroup

\textbf{Tasks}.
Six tasks (\autoref{tab:tasks}) were created to validate participants' ability to answer the questions (\autoref{question:knowledge} - \autoref{question:fail}) using RuleMatrix. For each task, we created two different questions with the same format (\eg, multiple-choice questions). That is, each participant was asked to perform 12 tasks. Questions of T1 to T5 were multiple-choice questions with one correct answer and four choices. 
T6(a) was also multiple-choice question while T6(b) asked the participants to enter a number.



\textbf{Results}. The average time that the participants took to complete all the 12 tasks in the formal study was 14' 43'' (std: 2' 26''). Accuracy of the performed tasks is summarized in \autoref{tab:tasks}. All the participants performed the required tasks fluently and correctly most of the time. This suggests validation of the basic usability of our method. However, we observed that participants took extra time in completing T2, which required the search and comparisons of multiple rules and multiple features. Three also complained that it was easy to get the wrong message from the textual representations provided in the choices in T1 and T2 (\ie, mistake \texttt{29 < x} from \texttt{x < 29}), and they had to double check to make sure that the clauses they identified in the visualization indeed matched the texts. We examined the answer sheets and found the errors of T1 are all of this type. This affirmed to us that text is not as intuitive as graphics in representing intervals in our context.

\textbf{Feedback}. We gathered feedback through subjective questionnaires after the formal study. \revision{Most participants felt that the supported interactions (expand, highlight and filter) are very ``useful and intuitive''. The detailed information provided by the data flow and support view was also regarded as ``helpful and just what they need''.} One participant liked how he could ``locate my hypotheses in the rules and understand how the model reacts, whether it is right or wrong, and how much observations in the dataset supports the hypotheses''. However, one participant had trouble understanding that there is only conjunctive relation between multiple clauses in a rule. Two participants suggested that a rule searching function would also be useful in validating hypotheses.




\section{Discussion and Conclusions}

In this work, we presented a technique for understanding classification models using rule-based explanations. We preliminarily validated the effectiveness of the rule induction algorithm on a set of benchmark datasets, and the effectiveness of the visual interface, RuleMatrix, through two use cases and a user study.

\textbf{Potential Usage Scenarios}. We anticipate the application of our method in domains where explainable intelligence is needed. Doctors can better utilize machine learning techniques for diagnosis and treatments with clear explanations. Banks can use efficient automatic credit approval systems while still being able to provide explanations to the applicants. Data scientists can better explain their findings when they need to present the results to no-experts. 

\textbf{Scalability of the Visualization}. Though the current implementation of the RuleMatrix can visualize rule lists with over 100 rules with over 30 features, the readability and understandability have only been validated on rule lists with less than 60 rules and 20 features. It is unclear whether users can still get an overall understanding of the model from such a complex list of rules. In addition, we used a qualitative color scheme to encode different classes. Though the effectiveness is limited to datasets with a limit number of classes, we assume that the method will be effective in most cases, since most classification tasks have fewer than 10 classes. It is also interesting to see if the proposed interface can be extended to support regression models by changing the qualitative color scheme to sequential color schemes.

\textbf{Scalability of the Rule Induction Method}. An intrinsic limitation of the rule induction algorithm results from the trade-off between the fidelity and complexity (interpretability) of the generated rule list. Depending on the complexity of the model and the domain, the algorithm would require a list containing hundreds of rules to approximate the model with an acceptable fidelity. The interpretability of rules also depends on the meaningfulness of the input features. This also limits the usage of our method in domains such as image classification or speech recognition. Another limitation is the current unavailability of efficient learning algorithms for rule lists. The SBRL algorithm takes about 30 minutes to generate a rule list from 200,000 samples and 14 features on server with 2.2GHz Intel Xeon. Its performance does not generalize well to datasets with an arbitrary number of classes. 

\textbf{Future Work}. One limitation of the presented work is that the method has not been fully validated with real experts in specific domains (\eg, health care). We expect to specialize the proposed method to meet the needs of specific domain problems (\eg, cancer diagnosis, or credit approvals) based on future collaborations with domain experts. Another interesting direction would be to systematically study the advantages and disadvantages of different knowledge representations (\eg, decision trees and rule sets) when considering human understandability. In other words, would people feel more comfortable with hierarchical representations (trees) or flat representations (lists) under different scenarios (\eg, verifying a prediction or understanding a complete model)? We regard this work as a preliminary and exploratory step towards explainable machine learning and plan to further extend and validate the idea of interpretability via inductive rules.






\bibliographystyle{abbrv-doi-hyperref-narrow}

\bibliography{template}

\begin{thebibliography}{10}
\renewcommand*{\sfdefault}{PTSansNarrow-TLF}

\bibitem{Abdul:2018:Trends}
\href{https://doi.org/10.1145/3173574.3174156}{A.~Abdul, J.~Vermeulen, D.~Wang,
  B.~Y. Lim, and M.~Kankanhalli}.
\newblock \href{https://doi.org/10.1145/3173574.3174156}{Trends and
  trajectories for explainable, accountable and intelligible systems: An hci
  research agenda}.
\newblock \href{https://doi.org/10.1145/3173574.3174156}{In {\em Proc.\ CHI
  Conference on Human Factors in Computing Systems}},
  \href{https://doi.org/10.1145/3173574.3174156}{pp. 582:1--582:18}.
  \href{https://doi.org/10.1145/3173574.3174156}{ACM},
  \href{https://doi.org/10.1145/3173574.3174156}{New York, NY, USA},
  \href{https://doi.org/10.1145/3173574.3174156}{2018}.
  \href{https://doi.org/10.1145/3173574.3174156}
{doi: \textsf{%
10\hspace{.1pt}\discretionary{.}{%
}{.}\hspace{.4pt}1145\discretionary{/}{%
}{/}3173574\hspace{.1pt}\discretionary{.}{%
}{.}\hspace{.4pt}3174156}}


\bibitem{Allahyari:2011:UserorientedAO}
H.~Allahyari and N.~Lavesson.
\newblock User-oriented assessment of classification model understandability.
\newblock In {\em Proc.\ 11th Conf.\ Artificial Inelligence}, 2011.

\bibitem{Andrews:1995:survey}
\href{https://doi.org/https://doi.org/10.1016/0950-7051(96)81920-4}{R.~Andrews,
  J.~Diederich, and A.~B. Tickle}.
\newblock
  \href{https://doi.org/https://doi.org/10.1016/0950-7051(96)81920-4}{Survey
  and critique of techniques for extracting rules from trained artificial
  neural networks}.
\newblock
  \href{https://doi.org/https://doi.org/10.1016/0950-7051(96)81920-4}{{\em
  Knowledge-Based Systems}},
  \href{https://doi.org/https://doi.org/10.1016/0950-7051(96)81920-4}{8(6):373
  -- 389},
  \href{https://doi.org/https://doi.org/10.1016/0950-7051(96)81920-4}{1995}.
  \href{https://doi.org/10.1016/0950-7051(96)81920-4}
{doi: \textsf{%
10\hspace{.1pt}\discretionary{.}{%
}{.}\hspace{.4pt}1016\discretionary{/}{%
}{/}0950\discretionary{%
}{-}{-}7051\discretionary{%
}{(}{(}96\discretionary{)}{%
}{)}81920\discretionary{%
}{-}{-}4}}


\bibitem{Augasta:2012:comparative}
\href{https://doi.org/10.1109/ICPRIME.2012.6208380}{M.~G. Augasta and
  T.~Kathirvalavakumar}.
\newblock \href{https://doi.org/10.1109/ICPRIME.2012.6208380}{Rule extraction
  from neural networks -- a comparative study}.
\newblock \href{https://doi.org/10.1109/ICPRIME.2012.6208380}{In {\em Proc.\
  Int.\ Conf.\ Pattern Recognition, Informatics and Medical Engineering
  (PRIME-2012)}}, \href{https://doi.org/10.1109/ICPRIME.2012.6208380}{pp.
  404--408}, \href{https://doi.org/10.1109/ICPRIME.2012.6208380}{Mar 2012}.
  \href{https://doi.org/10.1109/ICPRIME.2012.6208380}
{doi: \textsf{%
10\hspace{.1pt}\discretionary{.}{%
}{.}\hspace{.4pt}1109\discretionary{/}{%
}{/}ICPRIME\hspace{.1pt}\discretionary{.}{%
}{.}\hspace{.4pt}2012\hspace{.1pt}\discretionary{.}{%
}{.}\hspace{.4pt}6208380}}


\bibitem{Bilal:2018:cnnHierarchy}
\href{https://doi.org/10.1109/TVCG.2017.2744683}{A.~Bilal, A.~Jourabloo, M.~Ye,
  X.~Liu, and L.~Ren}.
\newblock \href{https://doi.org/10.1109/TVCG.2017.2744683}{Do convolutional
  neural networks learn class hierarchy?}
\newblock \href{https://doi.org/10.1109/TVCG.2017.2744683}{{\em IEEE
  Transactions on Visualization and Computer Graphics}},
  \href{https://doi.org/10.1109/TVCG.2017.2744683}{24(1):152--162},
  \href{https://doi.org/10.1109/TVCG.2017.2744683}{Jan 2018}.
  \href{https://doi.org/10.1109/TVCG.2017.2744683}
{doi: \textsf{%
10\hspace{.1pt}\discretionary{.}{%
}{.}\hspace{.4pt}1109\discretionary{/}{%
}{/}TVCG\hspace{.1pt}\discretionary{.}{%
}{.}\hspace{.4pt}2017\hspace{.1pt}\discretionary{.}{%
}{.}\hspace{.4pt}2744683}}


\bibitem{debock2010gam}
\href{https://doi.org/https://doi.org/10.1016/j.csda.2009.12.013}{K.~D. Bock,
  K.~Coussement, and D.~V. den Poel}.
\newblock
  \href{https://doi.org/https://doi.org/10.1016/j.csda.2009.12.013}{Ensemble
  classification based on generalized additive models}.
\newblock
  \href{https://doi.org/https://doi.org/10.1016/j.csda.2009.12.013}{{\em
  Computational Statistics \& Data Analysis}},
  \href{https://doi.org/https://doi.org/10.1016/j.csda.2009.12.013}{54(6):1535
  -- 1546},
  \href{https://doi.org/https://doi.org/10.1016/j.csda.2009.12.013}{2010}.
  \href{https://doi.org/10.1016/j.csda.2009.12.013}
{doi: \textsf{%
10\hspace{.1pt}\discretionary{.}{%
}{.}\hspace{.4pt}1016\discretionary{/}{%
}{/}j\hspace{.1pt}\discretionary{.}{%
}{.}\hspace{.4pt}csda\hspace{.1pt}\discretionary{.}{%
}{.}\hspace{.4pt}2009\hspace{.1pt}\discretionary{.}{%
}{.}\hspace{.4pt}12\hspace{.1pt}\discretionary{.}{%
}{.}\hspace{.4pt}013}}


\bibitem{breiman:1984:CART}
L.~Breiman, J.~Friedman, C.~J. Stone, and R.~A. Olshen.
\newblock {\em Classification and regression trees}.
\newblock CRC press, 1984.

\bibitem{Caruana:2015:IMH}
\href{https://doi.org/10.1145/2783258.2788613}{R.~Caruana, Y.~Lou, J.~Gehrke,
  P.~Koch, M.~Sturm, and N.~Elhadad}.
\newblock \href{https://doi.org/10.1145/2783258.2788613}{Intelligible models
  for healthcare: Predicting pneumonia risk and hospital 30-day readmission}.
\newblock \href{https://doi.org/10.1145/2783258.2788613}{In {\em Proc.\ 21th
  ACM SIGKDD Int.\ Conf.\ Knowledge Discovery and Data Mining}},
  \href{https://doi.org/10.1145/2783258.2788613}{KDD '15},
  \href{https://doi.org/10.1145/2783258.2788613}{pp. 1721--1730}.
  \href{https://doi.org/10.1145/2783258.2788613}{ACM},
  \href{https://doi.org/10.1145/2783258.2788613}{New York, NY, USA},
  \href{https://doi.org/10.1145/2783258.2788613}{2015}.
  \href{https://doi.org/10.1145/2783258.2788613}
{doi: \textsf{%
10\hspace{.1pt}\discretionary{.}{%
}{.}\hspace{.4pt}1145\discretionary{/}{%
}{/}2783258\hspace{.1pt}\discretionary{.}{%
}{.}\hspace{.4pt}2788613}}


\bibitem{Craven:1995:Trepan}
\href{http://dl.acm.org/citation.cfm?id=2998828.2998832}{M.~W. Craven and J.~W.
  Shavlik}.
\newblock \href{http://dl.acm.org/citation.cfm?id=2998828.2998832}{Extracting
  tree-structured representations of trained networks}.
\newblock \href{http://dl.acm.org/citation.cfm?id=2998828.2998832}{In {\em
  Proc.\ 8th Int.\ Conf.\ Neural Information Processing Systems}},
  \href{http://dl.acm.org/citation.cfm?id=2998828.2998832}{NIPS'95},
  \href{http://dl.acm.org/citation.cfm?id=2998828.2998832}{pp. 24--30}.
  \href{http://dl.acm.org/citation.cfm?id=2998828.2998832}{MIT Press},
  \href{http://dl.acm.org/citation.cfm?id=2998828.2998832}{Cambridge, MA, USA},
  \href{http://dl.acm.org/citation.cfm?id=2998828.2998832}{1995}.

\bibitem{uci:2017}
\href{http://archive.ics.uci.edu/ml}{D.~Dheeru and E.~Karra~Taniskidou}.
\newblock \href{http://archive.ics.uci.edu/ml}{{UCI} machine learning
  repository}, \href{http://archive.ics.uci.edu/ml}{2017}.

\bibitem{Fawcett:2008:PRIE}
\href{https://doi.org/10.1007/s10618-008-0089-y}{T.~Fawcett}.
\newblock \href{https://doi.org/10.1007/s10618-008-0089-y}{Prie: a system for
  generating rulelists to maximize roc performance}.
\newblock \href{https://doi.org/10.1007/s10618-008-0089-y}{{\em Data Mining and
  Knowledge Discovery}},
  \href{https://doi.org/10.1007/s10618-008-0089-y}{17(2):207--224},
  \href{https://doi.org/10.1007/s10618-008-0089-y}{Oct 2008}.
  \href{https://doi.org/10.1007/s10618-008-0089-y}
{doi: \textsf{%
10\hspace{.1pt}\discretionary{.}{%
}{.}\hspace{.4pt}1007\discretionary{/}{%
}{/}s10618\discretionary{%
}{-}{-}008\discretionary{%
}{-}{-}0089\discretionary{%
}{-}{-}y}}


\bibitem{Freitas:2014:position}
\href{https://doi.org/10.1145/2594473.2594475}{A.~A. Freitas}.
\newblock \href{https://doi.org/10.1145/2594473.2594475}{Comprehensible
  classification models: A position paper}.
\newblock \href{https://doi.org/10.1145/2594473.2594475}{{\em SIGKDD Explor.
  Newsl.}}, \href{https://doi.org/10.1145/2594473.2594475}{15(1):1--10},
  \href{https://doi.org/10.1145/2594473.2594475}{Mar 2014}.
  \href{https://doi.org/10.1145/2594473.2594475}
{doi: \textsf{%
10\hspace{.1pt}\discretionary{.}{%
}{.}\hspace{.4pt}1145\discretionary{/}{%
}{/}2594473\hspace{.1pt}\discretionary{.}{%
}{.}\hspace{.4pt}2594475}}


\bibitem{goodman2017a}
B.~Goodman and S.~Flaxman.
\newblock European union regulations on algorithmic decision-making and a
  "right to explanation".
\newblock {\em AI Magazine}, 38(3):50--57, 2017.

\bibitem{Han:2000:FPGrowth}
\href{https://doi.org/10.1145/335191.335372}{J.~Han, J.~Pei, and Y.~Yin}.
\newblock \href{https://doi.org/10.1145/335191.335372}{Mining frequent patterns
  without candidate generation}.
\newblock \href{https://doi.org/10.1145/335191.335372}{{\em ACM SIGMOD
  Record}}, \href{https://doi.org/10.1145/335191.335372}{29(2):1--12},
  \href{https://doi.org/10.1145/335191.335372}{May 2000}.
  \href{https://doi.org/10.1145/335191.335372}
{doi: \textsf{%
10\hspace{.1pt}\discretionary{.}{%
}{.}\hspace{.4pt}1145\discretionary{/}{%
}{/}335191\hspace{.1pt}\discretionary{.}{%
}{.}\hspace{.4pt}335372}}


\bibitem{Huysmans:2011:empirical}
\href{https://doi.org/https://doi.org/10.1016/j.dss.2010.12.003}{J.~Huysmans,
  K.~Dejaeger, C.~Mues, J.~Vanthienen, and B.~Baesens}.
\newblock \href{https://doi.org/https://doi.org/10.1016/j.dss.2010.12.003}{An
  empirical evaluation of the comprehensibility of decision table, tree and
  rule based predictive models}.
\newblock \href{https://doi.org/https://doi.org/10.1016/j.dss.2010.12.003}{{\em
  Decision Support Systems}},
  \href{https://doi.org/https://doi.org/10.1016/j.dss.2010.12.003}{51(1):141 --
  154}, \href{https://doi.org/https://doi.org/10.1016/j.dss.2010.12.003}{2011}.
  \href{https://doi.org/10.1016/j.dss.2010.12.003}
{doi: \textsf{%
10\hspace{.1pt}\discretionary{.}{%
}{.}\hspace{.4pt}1016\discretionary{/}{%
}{/}j\hspace{.1pt}\discretionary{.}{%
}{.}\hspace{.4pt}dss\hspace{.1pt}\discretionary{.}{%
}{.}\hspace{.4pt}2010\hspace{.1pt}\discretionary{.}{%
}{.}\hspace{.4pt}12\hspace{.1pt}\discretionary{.}{%
}{.}\hspace{.4pt}003}}


\bibitem{Kahng:2018:ActiVis}
\href{https://doi.org/10.1109/TVCG.2017.2744718}{M.~Kahng, P.~Y. Andrews,
  A.~Kalro, and D.~H.~. Chau}.
\newblock \href{https://doi.org/10.1109/TVCG.2017.2744718}{Activis: Visual
  exploration of industry-scale deep neural network models}.
\newblock \href{https://doi.org/10.1109/TVCG.2017.2744718}{{\em IEEE
  Transactions on Visualization and Computer Graphics}},
  \href{https://doi.org/10.1109/TVCG.2017.2744718}{24(1):88--97},
  \href{https://doi.org/10.1109/TVCG.2017.2744718}{Jan 2018}.
  \href{https://doi.org/10.1109/TVCG.2017.2744718}
{doi: \textsf{%
10\hspace{.1pt}\discretionary{.}{%
}{.}\hspace{.4pt}1109\discretionary{/}{%
}{/}TVCG\hspace{.1pt}\discretionary{.}{%
}{.}\hspace{.4pt}2017\hspace{.1pt}\discretionary{.}{%
}{.}\hspace{.4pt}2744718}}


\bibitem{Krause:2017:WVD}
J.~Krause, A.~Dasgupta, J.~Swartz, Y.~Aphinyanaphongs, and E.~Bertini.
\newblock A workflow for visual diagnostics of binary classifiers using
  instance-level explanations.
\newblock In {\em Proc.\ Visual Analytics Science and Technology (VAST)}. IEEE,
  Oct 2017.

\bibitem{Jan:2017:ASG}
\href{http://arxiv.org/abs/1711.09883}{J.~Leike, M.~Martic, V.~Krakovna, P.~A.
  Ortega, T.~Everitt, A.~Lefrancq, L.~Orseau, and S.~Legg}.
\newblock \href{http://arxiv.org/abs/1711.09883}{{AI} safety gridworlds}.
\newblock \href{http://arxiv.org/abs/1711.09883}{arXiv:1711.09883},
  \href{http://arxiv.org/abs/1711.09883}{2017}.

\bibitem{Letham:2015:BRL}
\href{https://doi.org/10.1214/15-AOAS848}{B.~Letham, C.~Rudin, T.~H. McCormick,
  and D.~Madigan}.
\newblock \href{https://doi.org/10.1214/15-AOAS848}{Interpretable classifiers
  using rules and bayesian analysis: Building a better stroke prediction
  model}.
\newblock \href{https://doi.org/10.1214/15-AOAS848}{{\em The Annals of Applied
  Statistics}}, \href{https://doi.org/10.1214/15-AOAS848}{9(3):1350--1371},
  \href{https://doi.org/10.1214/15-AOAS848}{Sep 2015}.
  \href{https://doi.org/10.1214/15-AOAS848}
{doi: \textsf{%
10\hspace{.1pt}\discretionary{.}{%
}{.}\hspace{.4pt}1214\discretionary{/}{%
}{/}15\discretionary{%
}{-}{-}AOAS848}}


\bibitem{Liu:2017:cnnvis}
\href{https://doi.org/10.1109/TVCG.2016.2598831}{M.~Liu, J.~Shi, Z.~Li, C.~Li,
  J.~Zhu, and S.~Liu}.
\newblock \href{https://doi.org/10.1109/TVCG.2016.2598831}{Towards better
  analysis of deep convolutional neural networks}.
\newblock \href{https://doi.org/10.1109/TVCG.2016.2598831}{{\em IEEE
  Transactions on Visualization and Computer Graphics}},
  \href{https://doi.org/10.1109/TVCG.2016.2598831}{23(1):91--100},
  \href{https://doi.org/10.1109/TVCG.2016.2598831}{Jan 2017}.
  \href{https://doi.org/10.1109/TVCG.2016.2598831}
{doi: \textsf{%
10\hspace{.1pt}\discretionary{.}{%
}{.}\hspace{.4pt}1109\discretionary{/}{%
}{/}TVCG\hspace{.1pt}\discretionary{.}{%
}{.}\hspace{.4pt}2016\hspace{.1pt}\discretionary{.}{%
}{.}\hspace{.4pt}2598831}}


\bibitem{Liu:2017:survey}
S.~Liu, X.~Wang, M.~Liu, and J.~Zhu.
\newblock Towards better analysis of machine learning models: A visual
  analytics perspective.
\newblock {\em Visual Informatics}, 1:48--56, 2017.

\bibitem{Marchand:2005:LDL}
M.~Marchand and M.~Sokolova.
\newblock Learning with decision lists of data-dependent features.
\newblock {\em Journal of Machine Learning Research}, 6:427--451, 2005.

\bibitem{Martens:2009:decomposeRule}
\href{https://doi.org/10.1109/TKDE.2008.131}{D.~Martens, B.~Baesens, and T.~V.
  Gestel}.
\newblock \href{https://doi.org/10.1109/TKDE.2008.131}{Decompositional rule
  extraction from support vector machines by active learning}.
\newblock \href{https://doi.org/10.1109/TKDE.2008.131}{{\em IEEE Transactions
  on Knowledge and Data Engineering}},
  \href{https://doi.org/10.1109/TKDE.2008.131}{21(2):178--191},
  \href{https://doi.org/10.1109/TKDE.2008.131}{Feb 2009}.
  \href{https://doi.org/10.1109/TKDE.2008.131}
{doi: \textsf{%
10\hspace{.1pt}\discretionary{.}{%
}{.}\hspace{.4pt}1109\discretionary{/}{%
}{/}TKDE\hspace{.1pt}\discretionary{.}{%
}{.}\hspace{.4pt}2008\hspace{.1pt}\discretionary{.}{%
}{.}\hspace{.4pt}131}}


\bibitem{Ming:2017:rnnvis}
Y.~Ming, S.~Cao, R.~Zhang, Z.~Li, Y.~Chen, Y.~Song, and H.~Qu.
\newblock Understanding hidden memories of recurrent neural networks.
\newblock In {\em Proc.\ Visual Analytics Science and Technology (VAST)}. IEEE,
  2017.

\bibitem{scikit-learn}
F.~Pedregosa, G.~Varoquaux, A.~Gramfort, V.~Michel, B.~Thirion, O.~Grisel,
  M.~Blondel, P.~Prettenhofer, R.~Weiss, V.~Dubourg, J.~Vanderplas, A.~Passos,
  D.~Cournapeau, M.~Brucher, M.~Perrot, and E.~Duchesnay.
\newblock Scikit-learn: Machine learning in {P}ython.
\newblock {\em Journal of Machine Learning Research}, 12:2825--2830, 2011.

\bibitem{Pezzotti:2018:deepeyes}
\href{https://doi.org/10.1109/TVCG.2017.2744358}{N.~Pezzotti, T.~Höllt, J.~V.
  Gemert, B.~P.~F. Lelieveldt, E.~Eisemann, and A.~Vilanova}.
\newblock \href{https://doi.org/10.1109/TVCG.2017.2744358}{Deepeyes:
  Progressive visual analytics for designing deep neural networks}.
\newblock \href{https://doi.org/10.1109/TVCG.2017.2744358}{{\em IEEE
  Transactions on Visualization and Computer Graphics}},
  \href{https://doi.org/10.1109/TVCG.2017.2744358}{24(1):98--108},
  \href{https://doi.org/10.1109/TVCG.2017.2744358}{Jan 2018}.
  \href{https://doi.org/10.1109/TVCG.2017.2744358}
{doi: \textsf{%
10\hspace{.1pt}\discretionary{.}{%
}{.}\hspace{.4pt}1109\discretionary{/}{%
}{/}TVCG\hspace{.1pt}\discretionary{.}{%
}{.}\hspace{.4pt}2017\hspace{.1pt}\discretionary{.}{%
}{.}\hspace{.4pt}2744358}}


\bibitem{Quinlan:1987:generateRule}
J.~R. Quinlan.
\newblock Generating production rules from decision trees.
\newblock In {\em Proc.\ 10th Int.\ Conf.\ Artificial Intelligence}, IJCAI'87,
  pp. 304--307. Morgan Kaufmann Publishers Inc., San Francisco, CA, USA, 1987.

\bibitem{Quinlan:1987:SDT}
\href{https://doi.org/10.1016/S0020-7373(87)80053-6}{J.~R. Quinlan}.
\newblock \href{https://doi.org/10.1016/S0020-7373(87)80053-6}{Simplifying
  decision trees}.
\newblock \href{https://doi.org/10.1016/S0020-7373(87)80053-6}{{\em
  International Journal of Human-Computer Studies}},
  \href{https://doi.org/10.1016/S0020-7373(87)80053-6}{27(3):221--234},
  \href{https://doi.org/10.1016/S0020-7373(87)80053-6}{Sep 1987}.
  \href{https://doi.org/10.1016/S0020-7373(87)80053-6}
{doi: \textsf{%
10\hspace{.1pt}\discretionary{.}{%
}{.}\hspace{.4pt}1016\discretionary{/}{%
}{/}S0020\discretionary{%
}{-}{-}7373\discretionary{%
}{(}{(}87\discretionary{)}{%
}{)}80053\discretionary{%
}{-}{-}6}}


\bibitem{Rauber:2017:project}
P.~E. Rauber, S.~G. Fadel, A.~X. Falcão, and A.~C. Telea.
\newblock Visualizing the hidden activity of artificial neural networks.
\newblock {\em IEEE Transactions on Visualization and Computer Graphics},
  23(1):101--110, Jan 2017.

\bibitem{ren:2016:squares}
\href{https://doi.org/10.1109/TVCG.2016.2598828}{D.~Ren, S.~Amershi, B.~Lee,
  J.~Suh, and J.~D. Williams}.
\newblock \href{https://doi.org/10.1109/TVCG.2016.2598828}{Squares: Supporting
  interactive performance analysis for multiclass classifiers}.
\newblock \href{https://doi.org/10.1109/TVCG.2016.2598828}{{\em IEEE
  Transactions on Visualization and Computer Graphics}},
  \href{https://doi.org/10.1109/TVCG.2016.2598828}{23(1):61--70},
  \href{https://doi.org/10.1109/TVCG.2016.2598828}{Jan 2017}.
  \href{https://doi.org/10.1109/TVCG.2016.2598828}
{doi: \textsf{%
10\hspace{.1pt}\discretionary{.}{%
}{.}\hspace{.4pt}1109\discretionary{/}{%
}{/}TVCG\hspace{.1pt}\discretionary{.}{%
}{.}\hspace{.4pt}2016\hspace{.1pt}\discretionary{.}{%
}{.}\hspace{.4pt}2598828}}


\bibitem{Ribeiro:2016:WIT}
\href{https://doi.org/10.1145/2939672.2939778}{M.~T. Ribeiro, S.~Singh, and
  C.~Guestrin}.
\newblock \href{https://doi.org/10.1145/2939672.2939778}{"{W}hy should {I}
  trust you?": Explaining the predictions of any classifier}.
\newblock \href{https://doi.org/10.1145/2939672.2939778}{In {\em Proc.\ 22nd
  ACM SIGKDD}}, \href{https://doi.org/10.1145/2939672.2939778}{KDD '16},
  \href{https://doi.org/10.1145/2939672.2939778}{pp. 1135--1144}.
  \href{https://doi.org/10.1145/2939672.2939778}{ACM},
  \href{https://doi.org/10.1145/2939672.2939778}{New York, NY, USA},
  \href{https://doi.org/10.1145/2939672.2939778}{2016}.
  \href{https://doi.org/10.1145/2939672.2939778}
{doi: \textsf{%
10\hspace{.1pt}\discretionary{.}{%
}{.}\hspace{.4pt}1145\discretionary{/}{%
}{/}2939672\hspace{.1pt}\discretionary{.}{%
}{.}\hspace{.4pt}2939778}}


\bibitem{Rissanen:1978:MDL}
\href{https://doi.org/https://doi.org/10.1016/0005-1098(78)90005-5}{J.~Rissanen}.
\newblock
  \href{https://doi.org/https://doi.org/10.1016/0005-1098(78)90005-5}{Modeling
  by shortest data description}.
\newblock
  \href{https://doi.org/https://doi.org/10.1016/0005-1098(78)90005-5}{{\em
  Automatica}},
  \href{https://doi.org/https://doi.org/10.1016/0005-1098(78)90005-5}{14(5):465
  -- 471},
  \href{https://doi.org/https://doi.org/10.1016/0005-1098(78)90005-5}{1978}.
  \href{https://doi.org/10.1016/0005-1098(78)90005-5}
{doi: \textsf{%
10\hspace{.1pt}\discretionary{.}{%
}{.}\hspace{.4pt}1016\discretionary{/}{%
}{/}0005\discretionary{%
}{-}{-}1098\discretionary{%
}{(}{(}78\discretionary{)}{%
}{)}90005\discretionary{%
}{-}{-}5}}


\bibitem{Rivest:1987:lists}
\href{https://doi.org/10.1023/A:1022607331053}{R.~L. Rivest}.
\newblock \href{https://doi.org/10.1023/A:1022607331053}{Learning decision
  lists}.
\newblock \href{https://doi.org/10.1023/A:1022607331053}{{\em Machine
  Learning}}, \href{https://doi.org/10.1023/A:1022607331053}{2(3):229--246},
  \href{https://doi.org/10.1023/A:1022607331053}{Nov 1987}.
  \href{https://doi.org/10.1023/A:1022607331053}
{doi: \textsf{%
10\hspace{.1pt}\discretionary{.}{%
}{.}\hspace{.4pt}1023\discretionary{/}{%
}{/}A\discretionary{:}{%
}{:}1022607331053}}


\bibitem{Sacha:2017:HumanCenteredMachineLearning}
\href{https://doi.org/10.1016/j.neucom.2017.01.105}{D.~Sacha, M.~Sedlmair,
  L.~Zhang, J.~A. Lee, J.~Peltonen, D.~Weiskopf, S.~C. North, and D.~A. Keim}.
\newblock \href{https://doi.org/10.1016/j.neucom.2017.01.105}{What you see is
  what you can change: Human-centered machine learning by interactive
  visualization}.
\newblock \href{https://doi.org/10.1016/j.neucom.2017.01.105}{{\em
  Neurocomputing}},
  \href{https://doi.org/10.1016/j.neucom.2017.01.105}{268(C):164--175},
  \href{https://doi.org/10.1016/j.neucom.2017.01.105}{Dec 2017}.
  \href{https://doi.org/10.1016/j.neucom.2017.01.105}
{doi: \textsf{%
10\hspace{.1pt}\discretionary{.}{%
}{.}\hspace{.4pt}1016\discretionary{/}{%
}{/}j\hspace{.1pt}\discretionary{.}{%
}{.}\hspace{.4pt}neucom\hspace{.1pt}\discretionary{.}{%
}{.}\hspace{.4pt}2017\hspace{.1pt}\discretionary{.}{%
}{.}\hspace{.4pt}01\hspace{.1pt}\discretionary{.}{%
}{.}\hspace{.4pt}105}}


\bibitem{schulz:2011:treevis}
\href{https://doi.org/10.1109/MCG.2011.103}{H.~J. Schulz}.
\newblock \href{https://doi.org/10.1109/MCG.2011.103}{Treevis.net: A tree
  visualization reference}.
\newblock \href{https://doi.org/10.1109/MCG.2011.103}{{\em IEEE Computer
  Graphics and Applications}},
  \href{https://doi.org/10.1109/MCG.2011.103}{31(6):11--15},
  \href{https://doi.org/10.1109/MCG.2011.103}{Nov 2011}.
  \href{https://doi.org/10.1109/MCG.2011.103}
{doi: \textsf{%
10\hspace{.1pt}\discretionary{.}{%
}{.}\hspace{.4pt}1109\discretionary{/}{%
}{/}MCG\hspace{.1pt}\discretionary{.}{%
}{.}\hspace{.4pt}2011\hspace{.1pt}\discretionary{.}{%
}{.}\hspace{.4pt}103}}


\bibitem{silverman:1986:density}
B.~W. Silverman.
\newblock {\em Density estimation for statistics and data analysis}, vol.~26.
\newblock CRC press, 1986.

\bibitem{Simonyan:2014:saliency}
K.~Simonyan, A.~Vedaldi, and A.~Zisserman.
\newblock Deep inside convolutional networks: Visualising image classification
  models and saliency maps.
\newblock In {\em Int.\ Conf.\ Learning Representations (ICLR) Workshop}, 2014.

\bibitem{smith1988pima}
J.~W. Smith, J.~Everhart, W.~Dickson, W.~Knowler, and R.~Johannes.
\newblock Using the adap learning algorithm to forecast the onset of diabetes
  mellitus.
\newblock In {\em Proc.\ Annu.\ Symp.\ Computer Application in Medical Care},
  p. 261. American Medical Informatics Association, 1988.

\bibitem{Strobelt:2018:lstmvis}
\href{https://doi.org/10.1109/TVCG.2017.2744158}{H.~Strobelt, S.~Gehrmann,
  H.~Pfister, and A.~M. Rush}.
\newblock \href{https://doi.org/10.1109/TVCG.2017.2744158}{Lstmvis: A tool for
  visual analysis of hidden state dynamics in recurrent neural networks}.
\newblock \href{https://doi.org/10.1109/TVCG.2017.2744158}{{\em IEEE
  Transactions on Visualization and Computer Graphics}},
  \href{https://doi.org/10.1109/TVCG.2017.2744158}{24(1):667--676},
  \href{https://doi.org/10.1109/TVCG.2017.2744158}{Jan 2018}.
  \href{https://doi.org/10.1109/TVCG.2017.2744158}
{doi: \textsf{%
10\hspace{.1pt}\discretionary{.}{%
}{.}\hspace{.4pt}1109\discretionary{/}{%
}{/}TVCG\hspace{.1pt}\discretionary{.}{%
}{.}\hspace{.4pt}2017\hspace{.1pt}\discretionary{.}{%
}{.}\hspace{.4pt}2744158}}


\bibitem{Tufte:2006:BE}
E.~R. Tufte.
\newblock {\em Beautiful Evidence}, chap.~2, pp. 46--63.
\newblock Graphis Pr, 2006.

\bibitem{tzeng:2005:nn}
F.-Y. Tzeng and K.-L. Ma.
\newblock Opening the black box-data driven visualization of neural networks.
\newblock In {\em Proc.\ Visualization}, pp. 383--390. IEEE, 2005.

\bibitem{elzen:2011:baobabview}
S.~van~den Elzen and J.~J. van Wijk.
\newblock Baobab{V}iew: Interactive construction and analysis of decision
  trees.
\newblock In {\em Proc.\ Visual Analytics Science and Technology (VAST)}, pp.
  151--160. IEEE, Oct 2011.

\bibitem{vanthienen:1994:decision}
J.~Vanthienen and G.~Wets.
\newblock From decision tables to expert system shells.
\newblock {\em Data \& Knowledge Engineering}, 13(3):265--282, 1994.

\bibitem{wang:2015:frl}
F.~Wang and C.~Rudin.
\newblock {Falling Rule Lists}.
\newblock In {\em Proc.\ 18th Int.\ Conf.\ Artificial Intelligence and
  Statistics}, vol.~38, pp. 1013--1022. PMLR, San Diego, California, USA, 2015.

\bibitem{Wongsuphasawat:2018:dataFlow}
\href{https://doi.org/10.1109/TVCG.2017.2744878}{K.~Wongsuphasawat, D.~Smilkov,
  J.~Wexler, J.~Wilson, D.~Mané, D.~Fritz, D.~Krishnan, F.~B. Viégas, and
  M.~Wattenberg}.
\newblock \href{https://doi.org/10.1109/TVCG.2017.2744878}{Visualizing dataflow
  graphs of deep learning models in tensorflow}.
\newblock \href{https://doi.org/10.1109/TVCG.2017.2744878}{{\em IEEE
  Transactions on Visualization and Computer Graphics}},
  \href{https://doi.org/10.1109/TVCG.2017.2744878}{24(1):1--12},
  \href{https://doi.org/10.1109/TVCG.2017.2744878}{Jan 2018}.
  \href{https://doi.org/10.1109/TVCG.2017.2744878}
{doi: \textsf{%
10\hspace{.1pt}\discretionary{.}{%
}{.}\hspace{.4pt}1109\discretionary{/}{%
}{/}TVCG\hspace{.1pt}\discretionary{.}{%
}{.}\hspace{.4pt}2017\hspace{.1pt}\discretionary{.}{%
}{.}\hspace{.4pt}2744878}}


\bibitem{Yang:2016:sbrl}
H.~Yang, C.~Rudin, and M.~Seltzer.
\newblock Scalable {B}ayesian rule lists.
\newblock In {\em Proc.\ 34th Int.\ Conf.\ Machine Learning {(ICML)}}, 2017.

\bibitem{zeiler14}
M.~D. Zeiler and R.~Fergus.
\newblock Visualizing and understanding convolutional networks.
\newblock In {\em ECCV}, pp. 818--833. Springer, Cham, 2014.

\end{thebibliography}

\clearpage
\begin{appendices}

\end{appendices}

\end{document}